\documentclass[preprint,12pt]{elsarticle}

\usepackage{amssymb}
\usepackage{amsthm}
\usepackage{amsmath}
\usepackage{xcolor}
\usepackage{placeins}
\usepackage{tabularx}
\usepackage{bm}
\usepackage{commath}
\usepackage{hyperref}
\biboptions{sort&compress}

\journal{Journal of Computational Physics}

\begin{document}

\begin{frontmatter}



\title{Physics informed cell representations for variational formulation of multiscale problems}


\author[inst1]{Yuxiang Gao}
\author[inst2]{Soheil Kolouri}
\author[inst1]{Ravindra Duddu}

\affiliation[inst1]{organization={Department of Civil and Environmental Engineering},
            addressline={Vanderbilt University}, 
            city={Nashville},
            postcode={37235}, 
            state={TN},
            country={United States of America}}

\affiliation[inst2]{organization={Department of Computer Science},
            addressline={Vanderbilt University}, 
            city={Nashville},
            postcode={37235}, 
            state={TN},
            country={United States of America}}

\begin{abstract}
With the rapid advancement of graphical processing units, Physics-Informed Neural Networks (PINNs) are emerging as a promising tool for solving partial differential equations (PDEs). However, PINNs are not well suited for solving PDEs with multiscale features, particularly suffering from slow convergence and poor accuracy. To address this limitation of PINNs, this article proposes physics-informed cell representations for resolving multiscale Poisson problems using a model architecture consisting of multilevel multiresolution grids coupled with a multilayer perceptron (MLP). The grid parameters (i.e., the level-dependent feature vectors) and the MLP parameters (i.e., the weights and biases) are determined using gradient-descent based optimization. The variational (weak) form based loss function accelerates computation by allowing the linear interpolation of feature vectors within grid cells. This cell-based MLP model also facilitates the use of a decoupled training scheme for Dirichlet boundary conditions and a parameter-sharing scheme for periodic boundary conditions, delivering superior accuracy compared to conventional PINNs. Furthermore, the numerical examples highlight improved speed and accuracy in solving PDEs with nonlinear or high-frequency boundary conditions and provide insights into hyperparameter selection. In essence, by cell-based MLP model along with the parallel tiny-cuda-nn library, our implementation improves convergence speed and numerical accuracy. 
\end{abstract}


\begin{highlights}
\item Developed a faster, more accurate cell-based MLP compared to PINNs for solving PDEs 
\item Grid structure in cell-based MLP accurately captures high-frequency solution features
\item Tiny-cuda-nn library for bilinear interpolation in grid cells reduces training time 
\item Proposed decoupled training scheme improves optimization with Dirichlet conditions 
\item Proposed parameter sharing scheme accurately enforces periodic boundary conditions
\end{highlights}


\end{frontmatter}


\section{Introduction}
\label{sec:introduction}
Multiscale Poisson problems are encountered in several computational physics applications, especially in the continuum modeling of porous and geological media. The partial differential equations (PDEs) corresponding to such problems typically describe reactive-diffusion phenomena that can occur at vastly different length scales due to high-frequency variations in diffusivity or reactivity. Developing powerful numerical methods to solve these equations while capturing features across multiple length scales is quite challenging and important in broader research areas of solid mechanics \cite{xu_multiscale_2021, qian_multiscale_2004}
, fluid mechanics \cite{abbaszadeh_reduced-order_2021,hoang_conservative_2019,kim_direct_2024}, and material design \cite{yu_phase_2022}. Conventional ways to tackle the multiscale PDEs involve refining the domain with a fine mesh \cite{xu_multiscale_2016} or applying asymptotic expansion with the finite element method (FEM) \cite{li_second-order_2016} or finite difference method (FDM) \cite{chen_high_2023}. With the rapid advancement of graphical processing units (GPU), machine learning (ML) frameworks have emerged as promising tools for solving complex PDEs involving highly nonlinear functions and intricate geometries \cite{asad_mechanics-informed_2023,taneja_multi-resolution_2023,park_convolution_2023,xiao_geometric_2023,liu_hidenn-fem_2023,xue_physics-embedded_2022,hernandez_port-metriplectic_2023,yang_using_2023,gao_cnn-based_2023,azizzadenesheli_neural_2024,geng_deep_2024,perera_graph_2022}. Among these tools, Physics-Informed Neural Networks (PINNs) stand out as one of the most promising options by obviating the need for training data generation and effortlessly incorporating physical laws into the learning process. The basic idea behind PINNs is to use a multilayer perceptron (MLP) that is optimized with a gradient-based algorithm to approximate the solution of the PDE, which is done while being supervised with physics-informed loss \cite{raissi_physics-informed_2019}.

Despite the success of PINNs in solving forward and inverse problems in many fields \cite{yuan_-pinn_2022,mowlavi_optimal_2023,gao_phygeonet_2021,chen_physics-informed_2021,lucor_simple_2022,khara_neural_2024, yang_multi-output_2022}, they often face challenges with slower convergence speeds and inaccurate solutions for PDEs with multiscale features, relative to FEM or FDM based implementations. One of the main reasons for this is the spectral bias \cite{rahaman_spectral_2019} of MLP neural networks, which struggle to approximate high-frequency features in the parameter field or solutions \cite{tancik2020fourier,moseley_finite_2023}. This is a known issue that has not been fully addressed inspite of considerable research efforts. Wang et al. \cite{wang_eigenvector_2021} showed that even with relatively simple problems, MLP networks struggled to approximate high-frequency features in the solutions of a 1D Poisson equation and a 1D Wave equation. 

Fourier feature embeddings have been proposed as a potential solution to the spectral representation challenges of MLPs \cite{tancik2020fourier,wang_eigenvector_2021}. However, the effectiveness of these embeddings can be limited as the embedded frequencies (e.g., the maximum frequency hyperparameter) are problem-dependent and necessitate fine-tuning for each distinct problem. Krishnapriyan et al. \cite{krishnapriyan_characterizing_2021} conducted an experiment that demonstrated that MLP networks have difficulty solving the 1-D convection equation with high-frequency features. To resolve this issue, they proposed a curriculum training scheme that solves PDEs in smaller time segments instead of the entire time domain to reduce the characteristic frequency of the PDE for solving. However, this method may not work well on high-frequency time-independent PDEs as the direction of information transfer is not clear. Leung et al.\cite{leung_nh-pinn_2022} showed that MLP networks also fail to solve the 1D elliptic equation with both high and low-frequency features. To resolve this issue, they applied the asymptotic expansion with homogenization to the multiscale PDE and solved the macro-scale and micro-scale PDE separately; however, this approach requires complex mathematical derivations, and the relative error is about 2\% compared to the ground truth FEM solution. 
Wang et al. \cite{wang_multi-stage_2024} developed multi-stage neural networks, where each training stage utilizes a new network optimized to fit the residual errors from the previous stage. This approach significantly reduced prediction errors, achieving near-machine-precision levels of accuracy. 

Another key challenge with using the standard PINN formulation arises from the incorporation of boundary conditions as weighted soft regularization terms in the loss function, which can cause numerical difficulties for optimization. Firstly, the weight factors assigned to different loss terms (i.e., the regularization hyper-parameters) can significantly affect the solution accuracy, which is demonstrated in Section \ref{sec:example}. Secondly, as discussed in Krishnapriyan et al. \cite{krishnapriyan_characterizing_2021}, soft regularization can lead to a complex loss landscape, making optimization cumbersome. Wang et al. \cite{wang_when_2022} developed an algorithm that adaptively updates weights by analyzing loss term convergence rates using the neural tangent kernel, achieving two orders of magnitude improvement in accuracy for 1-D wave equations with minimal extra computational cost. Goswami et al. \cite{goswami_transfer_2020} and Samaniego et al. \cite{samaniego_energy_2020} introduced a method to enforce Dirichlet boundary conditions automatically by adjusting the network output with a trial function, effective in complex solid mechanics PDEs, but challenging when dealing with several nonlinear boundary conditions. Schiassi et al. \cite{schiassi_extreme_2021}, applying the theory of connections framework \cite{mortari_multivariate_2019}, identified trial functions capable of meeting complex boundary conditions, though the approach requires case-specific mathematical derivations that can reduce network efficiency. Wang et al. \cite{wang_multi-stage_2024} optimized loss function weights by evaluating initial and pre-training loss ratios. Xie et al. \cite{xie_automatic_2023} utilized radial basis and neural functions to derive trial functions for any complex boundary condition, managing both Dirichlet and periodic boundaries in 2D and 4D PDEs with reasonable errors. Lastly, Leung et al. \cite{leung_nh-pinn_2022} proposed an oversampling strategy to boost network performance on periodic PDEs.

Grid-based structures combined with MLPs have become a popular model architecture in the field of computer vision and data compression \cite{takikawa_neural_2021,hadadan_neural_2021}. A model architecture proposed by Muller et al. \cite{muller_instant_2022} combined MLP with a multiresolution grid, and included hash encoding for data compression. Their studies showed that the model could accurately approximate complex 2D and 3D fields, including gigapixel images with small-scale details (i.e., high-frequency features) and neural radiance and density fields. Based on this prior work, we hypothesized that cell-based MLPs could be used to approximate the solution of PDEs with high-frequency features (i.e., coefficients or source terms) to improve accuracy and efficiency. The grid structure can be used to divide the model domain into smaller cells with multiple-resolution grids, which can capture localized features independently. Recently, Kang et al. \cite{kang_pixel_2023} solved linear and nonlinear PDEs with multi-level single-resolution cell-based MLP supervised by the strong form loss. Their model showed superior performance to standard PINNs with regard to accuracy and convergence speed for both forward and inverse problems, which supports our original hypothesis. 

In this article, we present multilevel multiresolution cell-based MLP supervised by weak form loss and leverage the grid structure to satisfy the Dirichlet BC and periodic BC, and demonstrate improved performance for resolving multiscale Poisson problems using diverse optimization schemes. We will first explain the model architecture, which contains a multilevel multiresolution grid integrated with MLP. The variational form-based loss function obviates the need for higher-order grid interpolation, thus accelerating computation. Applying spectral normalization to the MLP smooths the parameter optimization space and facilitates PDE solving. By utilizing the grid structure, we propose a decoupled training scheme for Dirichlet boundary conditions and a parameter sharing scheme for periodic boundary conditions, both delivering superior accuracy compared to the conventional PINNs. We present numerical examples involving multiscale Poisson problems with nonlinear boundary conditions or high-frequency features. To provide guidance on the selection of model hyperparameters for future applications, we further investigate how they affect the solution accuracy with the decoupled training and parameter sharing schemes.

\section{Methodology}
\label{sec:method}
\subsection{ Review of physics-informed neural networks}
We first briefly review the PINN formulation and introduce the notation by considering the following PDE form along with the boundary conditions:
\begin{equation}
    \mathcal{N}[u](\mathbf{x}) = f(\mathbf{x}), \ \ \mathbf{x}\in \Omega,
\end{equation}
\begin{equation}
    \mathcal{B}[u](\mathbf{x}) = g(\mathbf{x}), \ \ \mathbf{x}\in \partial \Omega,
\end{equation}
where $\mathcal{N}[\cdot]$ is a linear or non-linear differential operator and $\mathcal{B}[\cdot]$ is an operator for Dirichlet, Neumann, or periodic boundary conditions. The unknown scalar (solution) field denoted by $u(\mathbf{x})$ can be approximated using trainable neural networks $\psi(\mathbf{x}|\theta)$ with model parameters $\theta$. 

Following the strong form approach of Raissi et al. \cite{raissi_physics-informed_2019}, the approximation is achieved by minimizing the following physics-informed loss function:
\begin{equation}
    \mathcal{L}(\theta) =\lambda_{\mathrm{res}} \mathcal{L}_{\mathrm{res}}(\theta) +\lambda_{\mathrm{bc}} \mathcal{L}_{\mathrm{bc}}(\theta),
\end{equation}
where the residual loss is defined as
\begin{equation}
\label{eqn:residual-loss}
    \mathcal{L}_{\mathrm{res}}(\theta) = \frac{1}{N_{\mathrm{res}}}\sum_{r=1}^{N_{\mathrm{res}}}(\mathcal{N}[\psi](\mathbf{x}_r|\theta)-f(\mathbf{x}_r))^2,
\end{equation}
and the boundary condition loss is defined as
\begin{equation}
\label{eqn:PINN_BC}
    \mathcal{L}_{\mathrm{bc}}(\theta) = \frac{1}{N_{\mathrm{bc}}}\sum_{b=1}^{N_{\mathrm{bc}}}(\mathcal{B}[\psi](\mathbf{x}_b|\theta)-g(\mathbf{x}_b))^2.
\end{equation}
In the above equations, $N_{\mathrm{res}}$ and $N_{\mathrm{bc}}$ are the batch size of the sample points $\mathbf{x}_r$ and $\mathbf{x}_b$, and $\lambda_{\mathrm{res}}$ and $\lambda_{\mathrm{bc}}$ are the weight of different losses, which are usually user-specified. Standard gradient-based optimization algorithms (e.g., Adam and L-BFGS) are utilized to minimize the loss function, wherein the gradients with respect to input spatial coordinates $\mathbf{x}$ and model parameters $\theta$ are calculated by auto-differentiation.

An alternative approach to obtain the solution approximation is the energy form approach based on the Deep Ritz method proposed by E et al. \cite{e_deep_2018}. For example, an elliptic PDE can be written in its equivalent a variational or energy form:
\begin{equation}
    \min_{u\in H} E(u),
\end{equation}
\begin{equation}
    E(u) = \int_\Omega L(\mathbf{x},u(\mathbf{x}),\nabla u(\mathbf{x}))\ d\Omega,
\end{equation}
where $H$ is the set of admissible functions, $E(u)$ denotes the energy functional associated with the elliptic PDE, and $L$ is the corresponding Lagrangian operator that follows the Euler-Lagrange equation \cite{dacorogna_introduction_2014}. In the Deep Ritz method, the energy functional is evaluated using numerical integration
\begin{equation}
\label{eq:energy_loss}
    E(u(\theta))=\frac{1}{N_{\mathrm{E}}}\sum_{e=1}^{N_{\mathrm{E}}} L(\mathbf{x}_e,u(\mathbf{x}_e|\theta),\nabla u(\mathbf{x}_e|\theta)),
\end{equation}
where $N_{\mathrm{E}}$ is the batch size of sample points $\mathbf{x}_e$. In the variational form of the loss function, the Neumann boundary condition can be imposed as an external energy term with boundary integration, whereas imposing Dirichlet boundary conditions is not so straightforward because the deep neural network approximation functions generally do not satisfy the Kronecker delta property. A simple approach is to include the Dirichlet boundary conditions as penalty terms as follows \cite{e_deep_2018}:
\begin{equation}
    \mathcal{L}(\theta) = E(u|\theta) + \lambda_{\mathrm{DBC}} \mathcal{L}_{\mathrm{DBC}}(\theta)
    \label{eq:lossfunction}
\end{equation}
where $E(u|\theta)$ is the approximated energy function, $\mathcal{L}_\mathrm{DBC}(\theta)$ is the penalty term for the Dirichlet boundary condition defined in Eq. \eqref{eqn:PINN_BC}, and $\lambda_\mathrm{DBC}$ is the weight of the penalty term. To avoid choosing the penalty weight, Goswami et al. \cite{goswami_transfer_2020} imposed trial functions to the network to ensure the output satisfies the Dirichlet boundary condition automatically. However, as the trial function needs to be manually chosen, extending this method to the PDEs with nonlinear Dirichlet boundary conditions is difficult. To resolve this issue, we propose a decoupled training scheme by leveraging the architecture of the cell-based MLP model, which will be discussed in the following sections.

\subsection{Multi-level multiresolution cell-based MLP model}
\begin{figure}
    \centering
    \includegraphics[width=1.0\textwidth]{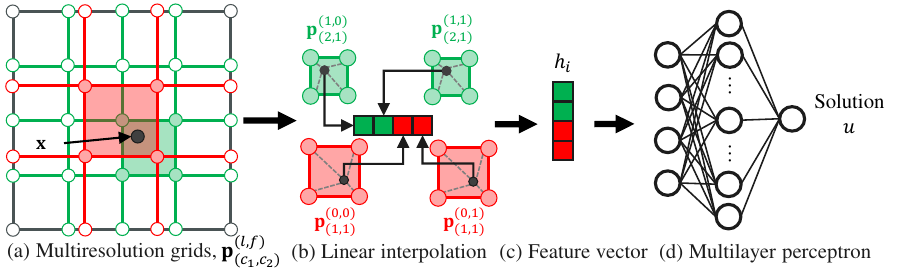}
    \caption{The model architecture of the multiresolution cell representations combined with MLPs, where the circles in the figure indicate trainable parameters. (a) The solution domain is divided into several levels of multiresolution grids with each grid further divided into cells and feature vectors (parameters) defined at the grid nodes. Given a sample point location, the model can determine the corresponding cell at all levels that contain the point; (b) In the cells containing the point at each level grid, the model then calculates the corresponding feature vector by linear interpolation from the nodal feature vectors; (c) The concatenated feature vector is passed to the MLP; (d) The MLP with its own trainable parameters predicts the solution variable by taking in the feature vector as input.}
    \label{fig:Model}
\end{figure}

The proposed model architecture comprises of multiresolution grids and MLPs, as shown in Figure \ref{fig:Model}. The model domain is divided into multiple levels of square grids with different grid resolutions, as defined by cell size. Vector parameters, which are the feature vectors defined at the grid nodes, can be updated by the optimizer based on the loss function. The size of the nodal feature vector $F$ is a hyperparameter taken equal to 2 in this study, but it can be taken as 4 or 8 depending on accuracy and efficiency constraints. Given a sample point location $\mathbf{x}$, the model algorithm can determine the corresponding cell at all levels that contain this point. The algorithm then calculates the corresponding feature vector at $\mathbf{x}$ using linear interpolation from the nodal feature vectors, similar to traditional grid-based numerical methods. Finally, the MLP has its own trainable model parameters and predicts the solution field $u(\mathbf{x})$ taking the feature vector as input. The detailed model formulations are discussed below in sections \ref{sec:grid_based_MLP_forward} and \ref{sec:grid_based_MLP_optimization}.
\FloatBarrier
\subsubsection{Forward calculation and the spacial gradient} \label{sec:grid_based_MLP_forward}
In this section, we present the details of the procedure for forward calculation given the sample point coordinates. 
Considering a multiresolution grid with a base (or minimum) resolution $r_\mathrm{base}$, number of levels $L$, number of features per level $F$, and the resolution ratio between the adjacent level $\beta$, the grid resolution of $l$-th level $r_l$ is determined as 
\begin{equation}
\label{eq:grid_resolution}
    r^{(l)} = \lfloor  \beta^{(l)}r_\mathrm{base} \rfloor, \ \ l = 0,1,2,...,L-1,
\end{equation}
where $\lfloor \cdot \rfloor$ denotes the floor or greatest integer function. Accordingly, for a 2D sample point $\mathbf{x}=[x_1, x_2] \in (0,1)^2$, the index vector $\mathbf{c}$ of the corresponding cell on the $l$-th level is 
\begin{equation}
    \mathbf{c} = [c_1,c_2] = \left[\lfloor x_1 r^{(l)} \rfloor, \lfloor x_2 r^{(l)} \rfloor\right].
\end{equation}
Then, the local coordinate of the sample point $\left[\xi^{(l)}_{1}, \xi^{(l)}_{2}\right]$ within the corresponding cell on level $l$ is calculated as
\begin{equation}
    \left[\xi^{(l)}_{1}, \xi^{(l)}_{2}\right] = \left[x_1 r^{(l)} - c_1, x_2 r^{(l)} - c_2\right],
\end{equation}
The feature vector elements $h_i$ at the sample point $\mathbf{x}$ are calculated from $l$-th level using bilinear interpolation as 
\begin{equation}
\label{eq:bilinear_interpolant}
\begin{split}
   h_i &= \begin{bmatrix} 1-\xi^{(l)}_{1} & \xi^{(l)}_{1} \end{bmatrix} \mathbf{p}^{(l,f)}_{(c_1,c_2)} \begin{bmatrix} 1-\xi^{(l)}_{2} \\ \xi^{(l)}_{2} \end{bmatrix}, \\
    \mathbf{p}^{(l,f)}_{(c_1,c_2)} &=  \begin{bmatrix} p_{(c_1,c_2)}^{(l,f)} & p_{(c_1,c_2+1)}^{(l,f)} \\ p_{(c_1+1,c_2)}^{(l,f)} & p_{(c_1+1,c_2+1)}^{(l,f)} \end{bmatrix} , \\
    i & = lF + f,\ \  f = 0,1,...,F-1,      
\end{split}
\end{equation}
where $\mathbf{p}^{(l,f)}_{(c_1,c_2)}$ is the trainable nodal parameter matrix for the $f$-th feature in the cell $[c_1,c_2]$ on $l$-th level of the multiresolution gird.

The MLP network projects the input feature vector at a sample point to the model output variable. Although the number of hidden layers of the MLP network can be arbitrary, for simplicity, the description below considers a single hidden layer network $\mathcal{M}$ with $W$ neurons in each layer and activation function $\phi$, so the output of the network is
\begin{equation} \label{eq:MLP}
    u = M\left(h_i\left(\mathbf{x}\right)\right) = \sum_{w=1}^{W} A_{w}^{(1)} \phi\left( \sum_{i=0}^{I}
    A^{(0)}_{wi}h_i\left(\mathbf{x}\right) + b_w^{(0)}\right)+b^{(1)},
\end{equation}
where $I+1=LF$ is the size of the feature vector $\mathbf{h}$, $\mathbf{A}^{(n)}$ are the trainable weight matrices of the $n$-th layer, and $\mathbf{b}^{(n)}$ are the trainable bias vectors of the $n$-th layer. 
The entire model $\mathcal{F}$ will output a scalar variable $u$ given a sample point and the corresponding spatial location $\mathbf{x}$ as
\begin{equation}
\label{eq:overall_formula}
    \mathcal{F}(\mathbf{x|p},\bm{\Theta})= \mathcal{M}\left( h_i(\mathbf{x|p})|\bm{\Theta}\right) = u,
\end{equation}
where $\bm{\Theta} = \{\mathbf{A}^{(n)},\mathbf{b}^{(n)}\}$ represent all model parameters of the MLP. 

For solving the Poisson equation with energy form loss (i.e., Eq. \eqref{eq:energy_loss}), we need to construct the loss function using the first-order spatial gradient calculated by auto-differentiation as
\begin{equation}\label{eq:spatial_gradient}
    \frac{\partial u}{\partial x_j} = \sum_{i=0}^{I} \frac{\partial u}{\partial h_i}\frac{\partial h_i}{\partial x_j}
\end{equation}
where the spatial index $j = \{1,2\}$ in 2D. The first term of Eq. \eqref{eq:spatial_gradient} is the derivative of the model output with respect to the feature vector given by:
\begin{equation}
\begin{split}
    \frac{\partial u}{\partial h_i}= \sum_{w=1}^W A^{(1)}_w\phi’\left(\sum_{i=0}^I A_{wi}^{(0)}h_i+b_w^{(0)}\right)A_{wi}^{(0)},
\end{split}
\end{equation}
where $\phi'$ is the derivative of the activation function $\phi$. The second term of Eq. \eqref{eq:spatial_gradient} is the local spatial gradient of the feature vector within the cell. Within a cell, as the index vector of the cell $\mathbf{c}$ and its nodal parameters $\mathbf{p}^{(l,f)}_{(c_1,c_2)}$ are constants, the local spatial gradient for index $i = lF + f$, is evaluated as:
\begin{equation}
\label{eq:spatial_gradient_cell}
\begin{split}
\frac{\partial h_i}{\partial x_j} &= \sum_{k=1}^{2} \frac{\partial h_i}{\partial x_k^{(l)}}\frac{\partial x_k^{(l)}}{\partial x_j}, \\
    \frac{\partial h_i}{\partial x_1^{(l)}} &= 
        \begin{bmatrix} -1 & 1 \end{bmatrix}  \mathbf{p}^{(l,f)}_{(c_1,c_2)} \begin{bmatrix} 1-\xi^{(l)}_{2} \\ \xi^{(l)}_{2} \end{bmatrix}, \\
    \frac{\partial h_i}{\partial x_2^{(l)}} &= 
        \begin{bmatrix} 1-\xi^{(l)}_{1} & \xi^{(l)}_{1} \end{bmatrix}  \mathbf{p}^{(l,f)}_{(c_1,c_2)} \begin{bmatrix} -1 \\ 1 \end{bmatrix}.    
\end{split}
\end{equation}
In the above equation, the derivative of the local coordinate with respect to the global coordinate is given by
\begin{equation}\label{eq:gradient_local_global}
  \frac{\partial x_k^{(l)}}{\partial x_j}  =
    \begin{cases}
      r^{(l)} , & \text{if } j = k,\\
      0, & \text{if } j \neq k.
    \end{cases} 
\end{equation}

\subsubsection{Model Optimization}\label{sec:grid_based_MLP_optimization}
In this section, we will discuss the parameter update scheme for grid parameters $\mathbf{p}$ and MLP parameters $\bm{\Theta}$ of the proposed model at a sample point $\mathbf{x}$ in 2-D. We will use the same notation as in Section \ref{sec:grid_based_MLP_forward}. The update scheme is performed using a first-order optimizer and uses the energy form loss $\mathcal{L}$ of an elliptical PDE. The loss is calculated by considering the model output $u$ and the first-order spatial gradient $\nabla u$. To balance conciseness with clarity, we use a combination of indicial and matrix notation to describe the parameter update scheme. 

During the training process, the optimizer updates the model parameters to approximate the solution to the PDE by minimizing the loss function. Most first-order optimization algorithms in machine learning are a variant of the gradient descent method. This method involves computing gradients and updating the parameters in the direction of the steepest descent. When training the cell-based MLP, the MLP parameters $\bm{\Theta}$ are updated as
\begin{equation}
    \bm{\Theta}^{(m+1)} = \bm{\Theta}^{(m)} - \gamma^{(m)}_{\bm{\Theta}} \nabla_{\bm{\Theta}} \left[\mathcal{L}\left(\mathbf{x,p,}\bm{\Theta}^{(i)}\right) \right],
\end{equation}
where $\mathcal{L}$ is the loss function calculated from the output of the model with parameters $\bm{\Theta}$ at iteration $m$, $\gamma^{(m)}_\theta$ is a hyperparameter of the optimization, also known as the learning rate, and $\nabla_{\bm{\Theta}} \left[\mathcal{L}(\bm{\Theta}^{(m)}) \right]$ is the Jacobian matrix that consists of the gradient of the loss with respect to the MLP parameter $\bm{\Theta}$.
The Jacobian matrix for the MLP parameters can be evaluated as follows:
\begin{equation}
\begin{split}\label{eq:Jacobian_theta}
    \nabla_{\bm{\Theta}} \left[\mathcal{L}(\bm{\Theta}) \right] &= 
    \frac{\partial}{\partial \bm{\Theta}}\mathcal{L}\left[\mathbf{x},u(\mathbf{x|p},\bm{\Theta}),\nabla u(\mathbf{x|p},\bm{\Theta}) \right] \\
    & = \frac{\partial \mathcal{L}}{\partial u}\frac{\partial u}{\partial \bm{\Theta}} + \frac{\partial \mathcal{L}}{\partial \nabla u}\frac{\partial \nabla u}{\partial \bm{\Theta}},\\
    & = \frac{\partial \mathcal{L}}{\partial u}\frac{\partial u}{\partial \bm{\Theta}} + \frac{\partial \mathcal{L}}{\partial \nabla u}\frac{\partial}{\partial \bm{\Theta}}\left(\frac{\partial u}{\partial \mathbf{h}}\right)\mathbf{\frac{\partial h}{\partial x}}.
\end{split}
\end{equation}
Except for $\mathbf{\frac{\partial h}{\partial x}}$, which is calculated using Eq. \eqref{eq:spatial_gradient_cell} and Eq. \eqref{eq:gradient_local_global}, all other terms in Eq. \eqref{eq:Jacobian_theta} are computed using PyTorch's auto-differentiation and the details are therefore not presented here in indicial notation. Similarly, the grid parameters $\mathbf{p}$ are updated as
\begin{equation}
    \mathbf{p}^{(m+1)} = \mathbf{p}^{(m)} - \gamma^{(m)}_{\mathbf{p}} \nabla_{\mathbf{p}} \left[\mathcal{L}\left(\mathbf{x,p}^{(m)},\bm{\Theta}\right) \right],
\end{equation}
where 
the Jacobian matrix for the grid parameters can be evaluated as:
\begin{equation}
\begin{split}\label{eq:Jacobian_p}
    \nabla_{\mathbf{p}} \left[\mathcal{L}(\mathbf{p}) \right] &= 
    \frac{\partial}{\partial \mathbf{p}}\mathcal{L}\left[\mathbf{x},\mathbf{u(x|p},\bm{\Theta}),\nabla\mathbf{u(x|p},\bm{\Theta}) \right] \\
    & = \frac{\partial \mathcal{L}}{\partial u}\frac{\partial u}{\partial \mathbf{p}} + \frac{\partial \mathcal{L}}{\partial \nabla u}\frac{\partial \nabla u}{\partial \mathbf{p}},\\
    & = \frac{\partial \mathcal{L}}{\partial u}\frac{\partial u}{\partial \mathbf{h}}\mathbf{\frac{\partial h}{\partial p}} + \frac{\partial \mathcal{L}}{\partial \nabla u}\left(\frac{\partial^2 u}{\partial \mathbf{h}^2}\mathbf{\frac{\partial h}{\partial p}\frac{\partial h}{\partial x}+\frac{\partial u}{\partial h}}\frac{\partial^2 \mathbf{h}}{\partial \mathbf{p \partial x}} \right).
\end{split}
\end{equation}
In the above Eq. \eqref{eq:Jacobian_p}, the partial derivative of the feature vector $\mathbf{h}$ with respect to the grid parameters $\mathbf{p}$, denoted as $\mathbf{\frac{\partial h}{\partial p}}$, is calculated for each level $l$ and feature component $f$:
\begin{equation}
\label{eq:dh_dp}
\begin{split}
\mathbf{\frac{\partial h}{\partial p}} &=  \left[ \frac{\partial h_i}{\partial p^{(l,f)}_{jk}} \right],\\
 \frac{\partial h_i}{\partial p^{(l,f)}_{jk}} &=
\begin{bmatrix} 1-\xi^{(l)}_{1} & \xi^{(l)}_{1} \end{bmatrix} 
\begin{bmatrix} \delta_{c_2k} \\ \delta_{(c_2+1)k} \end{bmatrix} 
\begin{bmatrix} \delta_{jc_1} & \delta_{j(c_1+1)} \end{bmatrix} 
\begin{bmatrix} 1-\xi^{(l)}_{2} \\ \xi^{(l)}_{2} \end{bmatrix},
\end{split}
\end{equation}
where $i = lF+f$ and $[c_1, c_2]$ are the cell indices on the $l$-th level containing the sample point $\mathbf{x}$. The partial derivative of the spatial gradient of the feature vector with respect to $\mathbf{p}$, denoted as $\frac{\partial^2 \mathbf{h}}{\partial \mathbf{p} \partial x}$, is evaluated as:
\begin{equation}
\label{eq:d2h_dpdx}
\begin{split}
\frac{\partial^2 \mathbf{h}}{\partial \mathbf{p \partial x}} &= \left[
\frac{\partial^2 h_i}{\partial p^{(l,f)}_{jk} \partial x_1},
\frac{\partial^2 h_i}{\partial p^{(l,f)}_{jk} \partial x_2}
\right],\\
\frac{\partial^2 h_i}{\partial p^{(l,f)}_{jk} \partial x_1}&= r^{(l)}
\begin{bmatrix} -1 & 1 \end{bmatrix} 
\begin{bmatrix} \delta_{c_2k} \\ \delta_{(c_2+1)k} \end{bmatrix} 
\begin{bmatrix} \delta_{jc_1} & \delta_{j(c_1+1)} \end{bmatrix} 
\begin{bmatrix} 1-\xi^{(l)}_{2} \\ \xi^{(l)}_{2} \end{bmatrix},\\
\frac{\partial^2 h_i}{\partial p^{(l,f)}_{jk} \partial x_2}&= r^{(l)}
\begin{bmatrix} 1-\xi^{(l)}_{1} & \xi^{(l)}_{1} \end{bmatrix} 
\begin{bmatrix} \delta_{c_2k} \\ \delta_{(c_2+1)k} \end{bmatrix} 
\begin{bmatrix} \delta_{jc_1} & \delta_{j(c_1+1)} \end{bmatrix} 
\begin{bmatrix} -1 \\ 1 \end{bmatrix},
\end{split}
\end{equation}
where the indices $j, k = 0,1,..., r_l-1$, and the Kronecker's delta is defined as
\begin{equation}\label{eq:Kronecker}
  \delta_{jk} =
    \begin{cases}
      1 , & \text{if } j = k,\\
      0, & \text{if } j \neq k.
    \end{cases} 
\end{equation}
All other terms in Eq. \eqref{eq:Jacobian_p} are computed using PyTorch’s auto-differentiation, and the details are therefore not presented here in indicial notation.

\subsection{Decoupled training scheme for Dirichlet boundary conditions} \label{sec:decoupled}

\FloatBarrier
\begin{figure}
    \centering
    \includegraphics[width=1.0\textwidth]{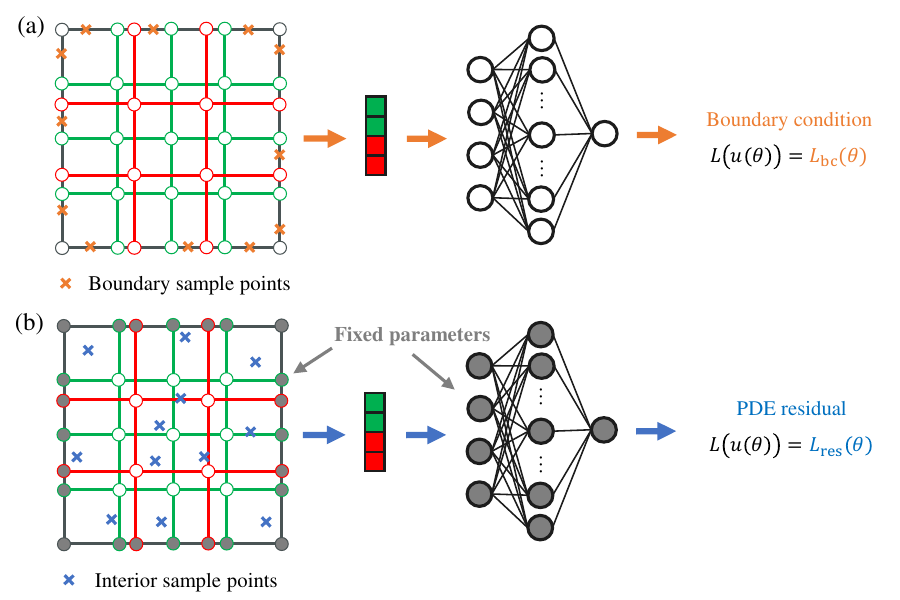}
    \caption{The decoupled training scheme: (a) In the first training phase, we disregard the PDE loss and solely focus on optimizing the model for boundary conditions by sampling the domain boundary. (b) In the second training phase, we freeze the parameters on the Dirichlet boundaries and MLPs (i.e., gray circles in the figure) and optimize the interior parameters with the PDE loss by sampling the interior domain.
}
    \label{fig:Decoupled}
\end{figure}

To obviate the need for selecting the weight factor $\lambda$ in the loss function in Eq. \eqref{eq:lossfunction}, we propose a decoupled training scheme designed specifically for Dirichlet boundary conditions. This is achieved by utilizing the architecture of multiresolution cell representation. Because the feature vectors at boundary sample points are determined using bilinear interpolation, they depend exclusively on the nodal feature vectors, which are model parameters on the boundaries. Therefore, only the nodal parameters on the boundaries and the parameters of the MLP influence the output function value on the boundary. Based on the above deduction, we propose the following two-phase training scheme to solve PDEs subject to Dirichlet boundary conditions:

\begin{enumerate}
    \item \textbf{First training phase:} We focus solely on sampling the domain boundary. The model is optimized to satisfy the Dirichlet boundary condition while disregarding the PDE loss. Once the training loss descends to an acceptable threshold, we freeze the nodal parameters associated with the Dirichlet boundaries and the MLP parameters. 
    \item \textbf{Second training phase:} We optimize the interior nodal parameters by sampling the interior domain and using the PDE loss. Because the boundary nodal and MLP parameters are frozen, the Dirichlet boundary condition is preserved during this second phase.
\end{enumerate}

The decoupled training scheme has several advantages. First, it eliminates the need for the weight factor $\lambda$ in Eq. \eqref{eq:lossfunction}. Second, it transforms a one-phase multi-task optimization problem into a two-phase single-task optimization process, improving the model's convergence rate and accuracy. Third, for enforcing Dirichlet boundary conditions on multiple boundaries, there are fewer parameters to update per training batch, which expedites the back propagation procedure and decreases training time. 
While the two phase scheme offers several benefits, it does come with certain limitations. Specifically, because the MLP is fixed during the second phase, the function approximation space provided by the MLP for the grid cells is not optimized based on the PDE loss. This means the MLP's universal approximation potential is not fully exploited. It is worth noting that a simpler function space can sometimes outperform a more complex one, and simply increasing the number of hidden layers in the MLP or adding more neurons to each layer might not necessarily lead to the expected improvement in solution accuracy.

\subsection{Parameter-sharing scheme for the periodic boundary conditions}
\label{sec:parameter-sharing}
\FloatBarrier
\begin{figure}
    \centering
    \includegraphics[width=1.0\textwidth]{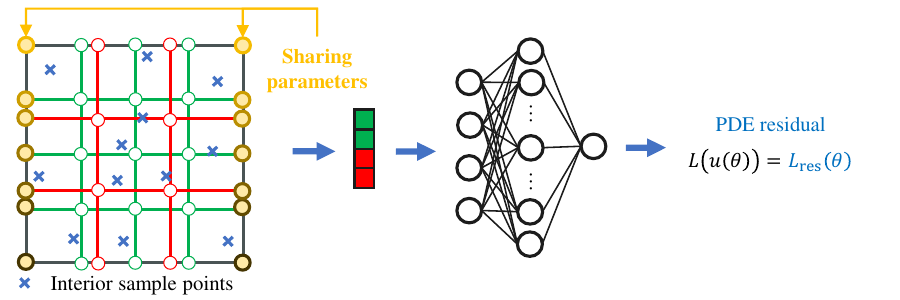}
    \caption{Schematic decription of the parameter sharing scheme. By sharing the nodal parameters on the grid that corresponds to the periodic boundary, the model's output always satisfies the appropriate conditions.
}
    \label{fig:parameter_sharing}
\end{figure}
In multiscale problems, such as those encountered in materials  with periodically varying features, periodic boundary conditions (PBCs) are often enforced on the boundaries of the repetitive unit cells or representative volume elements \cite{matous_review_2017}. To define the standard PINN loss that includes PBCs, it is necessary to choose the weight factor $\lambda$ and oversample the periodic boundary \cite{leung_nh-pinn_2022}. However, choosing an inappropriate weight factor can result in large solution error. To address this, we propose a parameter sharing scheme based on the model architecture with multiresolution grids. This novel yet simple scheme ensures that the model output satisfies the PBC by sharing the nodal parameters across the corresponding boundaries. 

As described in Section \ref{sec:decoupled}, the feature vectors at sample points are calculated using bilinear interpolation based on the nodal feature vectors, which are the nodal parameters. Thus, the feature vectors of the boundary sample points are solely determined by the nodal parameters on the respective boundaries. If the nodal parameters on two separate boundaries are the same, then the interpolated feature vectors and the subsequent MLP outputs will also be the same. Thus, parameter sharing between two boundaries ensures that the model output automatically satisfies the PBCs, so there is no need to include additional terms in the loss function to enforce PBCs.

The parameter sharing scheme offers similar benefits to the decoupled training scheme, described in Section \ref{sec:decoupled}. First, it eliminates the need for the weight factor $\lambda$. Second, it transforms a multi-task optimization problem into a single-task optimization problem, which improves the model's convergence rate and accuracy. Third, it reduces the number of parameters in the model, leading to less training time. As the model strictly adheres to PBCs, its accuracy is also improved.

\subsection{Spectral normalization}
The concept of ``smoothness" in a function can be defined through its Lipschitz continuity, which constrains the rate at which the function's values can change. This continuity for the MLP, denoted by $\mathcal{M}[\mathbf{h}]$, is guaranteed if the absolute value of the slope between any two feature vectors $\mathbf{h(x_1)}$ and $\mathbf{h(x_2)}$ does not exceed a limit, known as the Lipschitz constant $K$. Mathematically, this condition can be expressed as:
\begin{equation}
\frac{\norm{\mathcal{M}[\mathbf{h(x_1)}]-\mathcal{M}[\mathbf{h(x_2)}]}}{\norm{\mathbf{h(x_1)} - \mathbf{h(x_2)}}} \leq K,
\end{equation}
where $K \geq 0$ is a real number. Miyato et al. \cite{miyato_spectral_2018} introduced spectral normalization to control the Lipschitz continuity and improve training stability for generative adversarial networks (GANs). This procedure involves rescaling the MLP’s weight matrices $\mathbf{A}$ for all the subsequent MLP operations by their spectral norm $\sigma(\mathbf{A})$, which is the matrix's largest singular value:
\begin{equation}
\mathbf{A}_{\text{SN}} = \frac{\mathbf{A}}{\sigma(\mathbf{A})}.
\end{equation}
We found that when applying the decoupled training scheme for solving PDEs, spectral normalization can smooth the transformation from feature vector space to output values space performed by the MLP. This smoothing facilitates the optimization of cell parameters and aids in achieving solutions with reduced errors. For example, the spectral norm $\sigma(\mathbf{A})$ for the input or a hidden layer can be approximated using the power method, performed once per access to the weights in the training stage:
\begin{equation}
\sigma(\mathbf{A}) = \max_{\mathbf{h:h\neq0}}{ \frac{\norm{\mathbf{Ah}}_2}{\norm{\mathbf{h}}_2}}.
\end{equation}
The spectral normalized matrices $\mathbf{A}_{\text{SN}}$ are used in Eq. \eqref{eq:MLP} for all hidden layers. We implemented this spectral normalization procedure in PyTorch using the function
\texttt{torch.nn.utils.parametrizations.spectral\_norm}. 

\subsection{Model implementation}
The model implementation consists of two components: the multiresolution grids and MLPs.
For the multiresolution grid cells, there is no existing bilinear interpolation operator that supports back-propagation in PyTorch, so we need to develop PyTorch or CUDA code for the calculation from sample points $\mathbf{x}$ to the feature vector $\mathbf{h}$. It is a relatively simple method to fully use PyTorch to implement the operator for calculating the feature vector according to Eqs. \eqref{eq:grid_resolution}-- \eqref{eq:bilinear_interpolant}, because the automatic differentiation feature will calculate the necessary derivatives of the output with respect to the input and parameters. However, the PyTorch interpolation operator is typically more than an order of magnitude slower than the CUDA counterpart, though the latter requires the user to define the backward (gradient) calculation for the output with respect to the input (i.e., Eqs. \eqref{eq:spatial_gradient_cell}--\eqref{eq:gradient_local_global}) and the parameters (i.e., Eqs. \eqref{eq:dh_dp} -- \eqref{eq:d2h_dpdx}) in the CUDA language. 

To accelerate the proposed model for solving PDEs, we used the tiny-cuda-nn \cite{muller_tiny-cuda-nn_2021}, a lightning-fast C++/CUDA neural network framework, to implement the bilinear interpolation in multiresolution grids. 
We modified the bilinear interpolation cell kernel in tiny-cuda-nn to apply the decoupled training scheme and the parameter sharing scheme described in Sections \ref{sec:decoupled} and \ref{sec:parameter-sharing}, respectively. This modification aligns the boundaries of all levels of grids with the domain boundaries and assigns trainable parameter groups based on the locations, in accordance with the domain boundaries. Our changes ensure that the boundary output values depend solely on the grid parameters at all boundary levels and can be optimized separately.
Finally, we integrate our PyTorch-based MLP module with the CUDA-based bilinear interpolation cell to obtain the full model implementation (see the statement on data availability).

\subsection{Error metric}
For the example cases where an analytical solution to the PDE exists, we compare our model results with the solution at $N\times N$ grid points. However, for the cases where the analytical solution to the PDE cannot be found, we discretize the domain into a $(N-1)\times (N-1)$ structured triangle mesh in FEniCS software and compare our model results with the numerical solution from the FEM at the mesh nodes. We measure the accuracy of the ML model predictions against the ground truth (either the FEM or analytical solution) using the normalized root mean square error (NRMSE) defined as: 
\begin{equation}
\begin{split}
\text{NRMSE}
= \sqrt{\dfrac{\sum_{i=1}^{N^2} (Y_{i} - \Bar{Y}_{i})^2}{\sum_{i=1}^{N^2} Y_{i}^2}},
\end{split}
\label{eq:nrmse}
\end{equation}
where $Y$ represents the ground truth values and $\Bar{Y}$ represents our cell-based MLP model solution on the evaluation points.

\section{Numerical examples}
\label{sec:example}
In this section, we compare the solution accuracy and training time with other PINN models in the literature and a single-level grid without MLP by solving four different PDEs with the optimization scheme we proposed in Section \ref{sec:method}. The single-level grid without MLP serves as a baseline model, which uses with gradient descent-based optimization methods and CUDA acceleration for solving PDEs. We have chosen a grid resolution of 90 for the single-level baseline model, closely matching the maximum resolution of the multiresolution grid in the cell-based MLP, to compare model performance.  All standard PINNs are five-layer MLPs with 64 neurons per layer and Tanh as the activation function. The hyperparameters of the cell-based MLP are listed in Table \ref{tab:Exp_hyperparameters}. For the sake of consistency, all models were trained using the Adam optimizer \cite{kingma2014adam} with an initial learning rate of 0.005. The learning rate decays at a rate of 0.4 after every 800 steps, for a total of 6000 optimization steps. For each step, we randomly take 2000 sample points from the boundary to calculate the boundary condition loss and 30000 sample points from the interior domain to calculate the PDE loss. We repeat each experiment 10 times to prevent the results from being significantly affected by special random seeds. The maximum, minimum, and mean errors of the ten experiments are reported in each plot. In sections \ref{sec:Exp2} and \ref{sec:Exp3}, because there are no well-known analytical solutions, we solved the PDEs using FEniCS \cite{alnaes_fenics_2015}. We used structured triangle meshes for the unit square domain with a characteristic mesh size of 0.002, which is adequately fine to provide accurate solutions. We calculated the error between the FEM nodal solution and the corresponding solution obtained from the PINN or cell-based MLP models.

\begin{table}[h!] 
\centering
\begin{tabular}{ >{\raggedright\arraybackslash}p{7cm}  >{\raggedright\arraybackslash}p{5cm}} 
\hline
Hyperparameter & Value \\
\hline
Number of grid levels ($L$) & 16  \\
Max cell resolution ($r_\text{max}$)  & 87  \\
Resolution ratio of adjacent levels ($\beta$)   & 1.12\\
Number of features per level ($F$) & 2\\
Number of MLP hidden layers & 1 \\
Number of MLP neurons per layer & 32\\
Activation function of MLP & sin\\
\hline
\end{tabular}
\caption{Hyperparameters of the multiresolution cell-based MLP}
\label{tab:Exp_hyperparameters}
\end{table}
\FloatBarrier
\subsection{Poisson equation with non-constant Dirichlet BC} \label{sec:Exp1}
We consider a 2D Poisson equation with non-constant Dirichlet BC imposed on four boundaries solved in Lagaris et al.\cite{lagaris_artificial_1998} and Schiassi et al.\cite{schiassi_extreme_2021}. This example aims to demonstrate the numerical accuracy and superior convergence rate of the decoupled training scheme relative to the standard (penalty-based) coupled training scheme for solving PDEs with nonlinear Dirichlet boundary conditions. The strong form is 
\begin{align}
\begin{split}\label{eq:Poisson_DBC_strong}
    \nabla u &= f(x,y),\\
    f(x,y) &= e^{-x}\left(x-2+y^3+6y\right),
\end{split}    
\end{align}
where $(x,y)\in \Omega = [0,1]^2$, and the associated boundary conditions are
\begin{align}
\begin{split}
    u(0,y) &= y^3,\\
    u(1,y) &= \left(1+y^3\right)e^{-1},\\
    u(x,0) &= xe^{-x},\\
    u(x,1) &= e^{-x}(x+1).
\end{split}
\end{align}
The true solution is
\begin{equation}
\label{eq:Exp1_true}
    u(x,y) = e^{-x}\left(x+y^3\right)
\end{equation}
The variational energy form of Eq. \eqref{eq:Poisson_DBC_strong} is
\begin{equation}
    E = \int_\Omega \left[\frac{1}{2}\left( \nabla u\right)^2 + fu \right] \, d\Omega
\end{equation}

\begin{figure}
    \centering
    \includegraphics[width=1.0\textwidth]{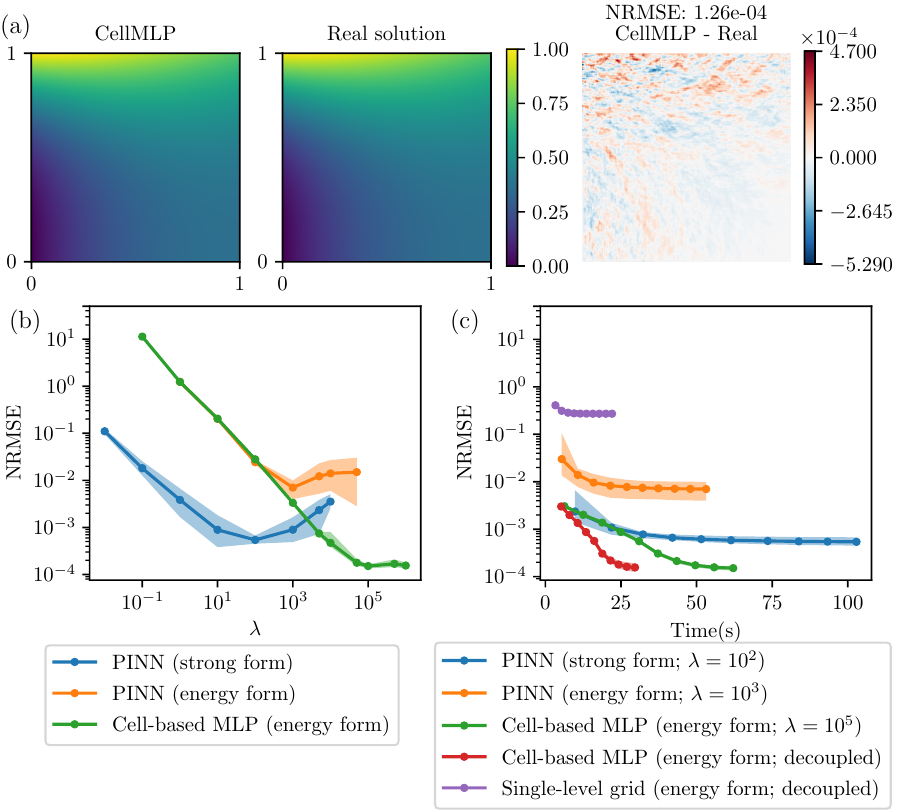}
    \caption{Model performance evaluation for the Poisson equation with non-constant Dirichlet BC. (a) Solution of cell-based MLP with decoupled training scheme, analytical solution, and the difference between them; (b) The relative error (NRMSE) of the final solutions for different weight factor $\lambda$ in loss function for different models; (c) The relative error (NRMSE) during the training process for different models. }
    \label{fig:Exp1_results}
\end{figure}

As shown in Figure \ref{fig:Exp1_results}(a), the solution from the cell-based MLP with decoupled training scheme matches with the analytical solution (i.e., Eq. \eqref{eq:Exp1_true}), with a NRMSE less than 0.0145 \%. Figure \ref{fig:Exp1_results}(b) shows the weight factor $\lambda$ will significantly affect the solution error for the standard PINN model and the cell-based MLP model, when enforcing the Dirichlet boundary condition by adding a penalty term in the loss function. For example, for $\lambda=1$ the NMRSE is greater than 100 \%, but with the right choice of $\lambda=10^5$ the NMRSE reduces to 0.0145 \%. 

As shown in Figure \ref{fig:Exp1_results}(c), the cell-based MLP trained with the decoupled training scheme converges faster and achieves higher accuracy compared to other models. Even for the optimal value of $\lambda\approx 10^2$ from Figure \ref{fig:Exp1_results}(b), the final error of the standard (strong/energy form) PINN is greater than the cell-based MLP, which indicates that the latter is relatively a better representation. It is worth noting that the single-level grid has a greater final error compared to the cell-based MLP, which suggests that the lower-resolution grids and the MLP are crucial for solving the Poisson equation with highly nonlinear Dirichlet BCs. 
If we choose the optimal value of $\lambda$, the cell-based MLP model with and without the decoupled training scheme will achieve a similar level of accuracy, given the same number of optimization steps. However, each optimization step will take longer without the decoupled training scheme, because it reduces the number of parameters that need to be optimized by fixing some of the parameters in the second training phase. Consequently, a smaller Jacobian matrix needs to be calculated for the gradient descent. Thus, this study shows that the decoupled training scheme provides a significant advantage over the coupled training scheme. 

\FloatBarrier

\subsection{Poisson equation with high-frequency variable coefficient} \label{sec:Exp2}
Flow in porous media (e.g., groundwater systems, firn aquifers in glaciers) is described by Poisson equations based on Darcy's flow low. Often, these equations are difficult to solve using standard PINNs to achieve high-frequency variability in the permeability of the porous media. The purpose of this example is to demonstrate the advantage of using cell-based representations to resolve these so-called multiscale Poisson problems. We consider a high-frequency variable coefficient Poisson equation discussed in Leung et al. \cite{leung_nh-pinn_2022}. The strong form is 
\begin{equation}
\begin{split}\label{eq:var_cof_Poisson}
    -\nabla \cdot \left(a(x,y) \nabla u \right) &= f(x,y),\\
    a(x,y) &= 2 + \text{sin}\left( \frac{2\pi x}{\epsilon} \right) \text{cos}\left( \frac{2 \pi y}{\epsilon} \right),\\
    f(x,y) &= \text{sin}(x) + \text{cos}(y),
\end{split}
\end{equation}
where $(x,y)\in \Omega = [0,1]^2$, and the value of $\epsilon$ in the permeability term $a(x,y)$ has an impact on the frequency of spatial variation in permeability. The associated boundary condition is
\begin{equation}
    u(x,y) = 0, \ \ (x,y) \in \partial \Omega.
\end{equation}
The variational energy form of Eq. \eqref{eq:var_cof_Poisson} is
\begin{equation}
    E = \int_\Omega \left(\frac{1}{2}a(\nabla u)^2 - fu\right)d\Omega.
\end{equation}

\begin{figure}
    \centering
    \includegraphics[width=1.0\textwidth]{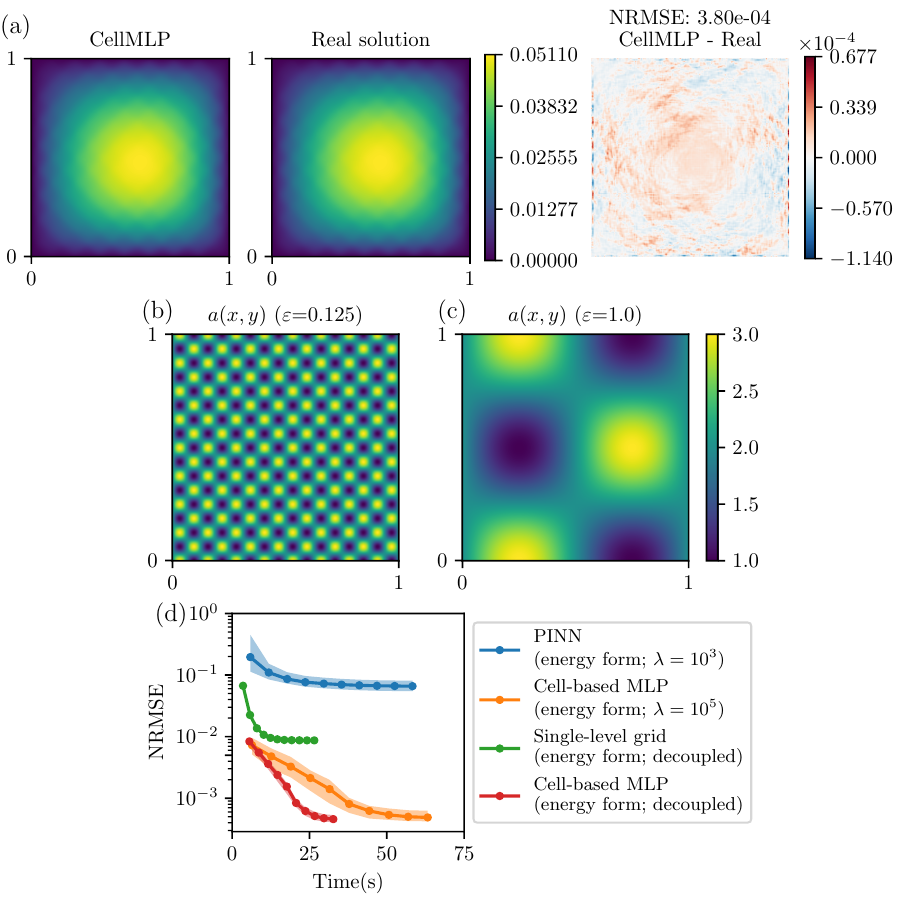}
    \caption{Model performance evaluation and equation properties for Poisson equation with high-frequency variable coefficient: (a) Solution of cell-based MLP with decoupled training scheme, FEM solution, and the difference between them. The permeability field $a(x,y)$ with (b) $\varepsilon$ = 0.125 and (c) $\varepsilon$ = 1. (d) The relative error (NRMSE) during the training process for different models. }
    \label{fig:Exp2_results}
\end{figure}

As shown in Figure \ref{fig:Exp2_results}(a), the solution from the cell-based MLP with decoupled training scheme matches with the solution obtained from the finite element method (FEM), with the NRMSE less than 0.0386 \%. As there is no known analytical solution, we use the FEM solution as the ground truth, with a mesh resolution comparable to the cell size of the base resolution grid. As shown in Figures \ref{fig:Exp2_results}(b) and (c), a smaller $\epsilon$ results in higher frequency variations in permeability, which makes the governing PDE more difficult to solve using the standard PINN.
To keep consistent with Leung et al. and demonstrate the advantage of the cell-based MLP model, we take $\varepsilon = 1/8$.


\FloatBarrier

\begin{table}[h!] 
\centering
\begin{tabular}{ >{\raggedright\arraybackslash}p{6.5cm}  >{\raggedright\arraybackslash}p{2cm}  >{\raggedright\arraybackslash}p{3.5cm} }
\hline
Model & NRMSE & Training time (s) \\
\hline
Grid MLP with decoupled training & $4.3\times 10^{-4}$ & 33 \\

Grid MLP with energy form loss  & $4.8\times 10^{-4}$ & 63 \\

Single-level grid  & $8\times 10^{-3}$ & 28 \\

Energy form PINN  & 0.0659 & 58 \\

Asymptotic expansion with PINN\cite{leung_nh-pinn_2022}  & 0.0213 & -- \\

Strong form PINN\cite{leung_nh-pinn_2022}  & 0.909 & -- \\
\hline
\end{tabular}
\caption{The NRMSE and the training time of different models for solving the high-frequency variable coefficient Poisson equation.}
\label{tab:Exp2}
\end{table}

\FloatBarrier
As shown in Figure \ref{fig:Exp2_results}(d), similar to the results in Section \ref{sec:Exp1}, the cell-based MLP trained with the decoupled training scheme converges faster and achieves higher accuracy compared to other models. Here, we observe a behavior similar to that described in Section \ref{sec:Exp1}: the cell-based MLP can also yield a low-error solution when incorporating the boundary condition loss as a penalty term with an optimal weight factor; however, it requires longer training time compared to the decoupled training scheme. The final error of the single-level grid using the decoupled training scheme is lesser than the one obtained through the PINN with energy form loss, but greater than the cell-based MLP. This suggests that the high-resolution grid is useful for capturing high-frequency features in the PDE solution, whereas the lower-resolution grid and the MLP are useful for capturing macroscale low-frequency features.
As shown in Table \ref{tab:Exp2}, the cell-based MLP with decoupled training is almost two orders of magnitude more accurate than the method proposed in Leung et al. \cite{leung_nh-pinn_2022} and three orders of magnitude more accurate than the standard strong form PINN. We are unable to compare the training time because Leung et al. \cite{leung_nh-pinn_2022} didn't report the training time; also, it is machine (GPU) dependent, so it is not meaningful to compare our time with theirs. 

\subsection{Variable coefficient Poisson equation with periodic boundary conditions}\label{sec:Exp3}
For multiscale problems, direct numerical simulation is not viable, so it is beneficial to use consider a coarse/macro scale and fine/micro scale split of the problem. The representative volume element (RVE) or unit cell is often used to define the fine scale problem domain and often periodic boundary conditions are applied to enforce the repetitive nature of the RVE at the coarse scale. The purpose of this example is to demonstrate the parameter sharing scheme proposed for the cell-based representations offers a distinct advantage for enforcing periodic boundary conditions, which is otherwise problematic for standard PINNs using MLPs. We consider a variable coefficient Poisson equation with periodic boundary conditions, which is the unit cell problem of the PDE solved in section \ref{sec:Exp2} with asymptotic expansion. Hence, the strong form is: 
\begin{equation}
\begin{split}\label{eq:PBC_var_cof_Poisson}
    -\nabla \cdot \left(a(x,y) \nabla u \right) &= f(x,y),\\
    a(x,y) &= 2 + \text{sin}\left( 2\pi x \right) \text{cos}\left( 2 \pi y \right),\\
    f(x,y) &= 2\pi\text{cos}(2\pi x)\text{cos}(2\pi y),
\end{split}
\end{equation}
where $(x,y)\in \Omega = [0,1]^2$, the periodic boundary conditions are imposed on all boundaries, and the four corners are fixed.
\begin{equation}
\begin{split}
u(0,y) - u(1,y)&=0,\\
u(x,0) - u(x,1)&=0,\\
u(0,0) &= 0.
\end{split}
\end{equation}

\begin{figure}
    \centering
    \includegraphics[width=1.0\textwidth]{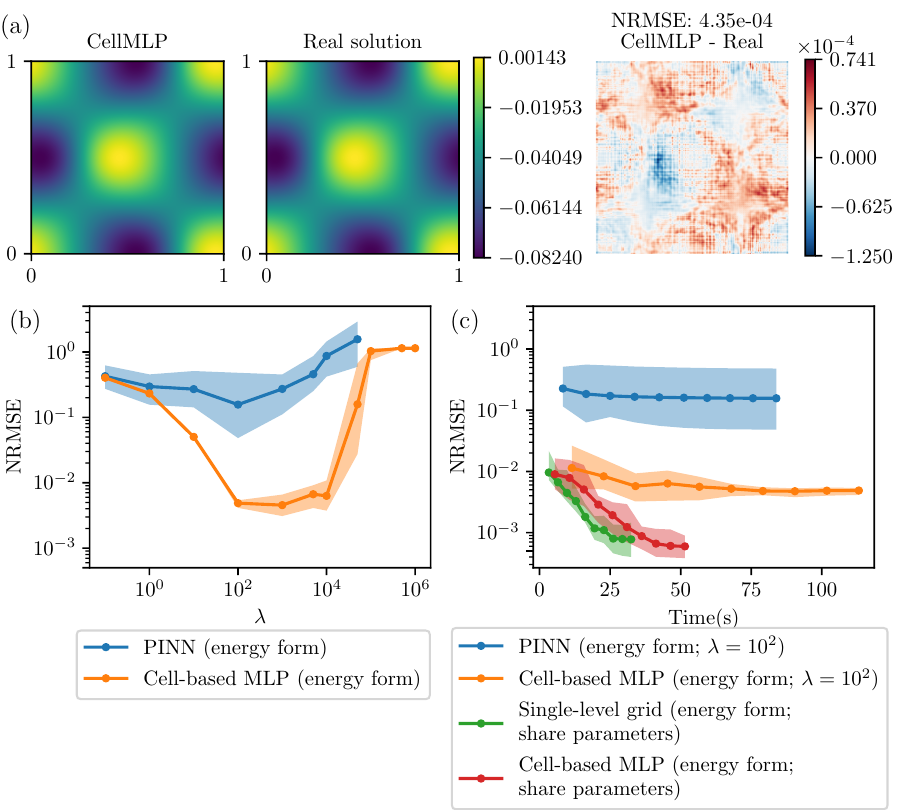}
    \caption{Model performance evaluation for the Variable coefficient Poisson equation with PBC. (a) Solution of cell-based MLP with parameter sharing scheme, FEM solution, and the difference between them; (b) The relative error (NRMSE) of the final solutions for different weight factor $\lambda$ in loss function for different models; (c) The relative error (NRMSE) during the training process for different models. }
    \label{fig:Exp3_results}
\end{figure}

As shown in Figure \ref{fig:Exp3_results}(a), the solution from the cell-based MLP with the parameter sharing scheme matches with the FEM solution, with the NRMSE less than 0.056 \%.
Figure \ref{fig:Exp3_results}(b) shows the weight factor $\lambda$ will significantly affect the solution error for the standard PINN and the cell-based MLP models for enforcing the periodic boundary condition terms in the loss function. If $\lambda$ is not chosen appropriately then the error can be as high as 100 \%.

\FloatBarrier
As shown in Figure \ref{fig:Exp3_results}(c), for the optimal value of $\lambda=10^2$ (see Fig. \ref{fig:Exp3_results}b) with the periodic boundary condition loss, the solution error of the cell-based MLP is lesser than the standard PINN. When utilizing the parameter sharing scheme, the cell-based MLP and single-level grid implementations converge faster and produce a more accurate final solution than the other two implementations. The single-level grid's calculation time for each optimization step is shorter due to its simpler model structure for back-propagation algorithms. This study indicates that the cell-based MLP is beneficial for solving unit cell problems with periodic boundary conditions. 
Additionally, the parameter sharing scheme improves accuracy and convergence rate by converting a multi-task optimization problem to a single-task problem and reduces the number of parameters that need to be updated to enforce the boundary condition. 

\begin{table}[h!] 
\centering
\begin{tabular}{ >{\raggedright\arraybackslash}p{7.6cm}  >{\raggedright\arraybackslash}p{1.7cm}  >{\raggedright\arraybackslash}p{3.1cm} }
\hline
Model & NRMSE & Training time (s) \\
\hline
Grid MLP with parameter sharing & $6.0\times 10^{-4}$ & 51 \\

Grid MLP with energy form loss  & $4.9\times 10^{-3}$ & 113 \\

Single-level grid  & $7.7\times 10^{-4}$ & 35 \\

Energy form PINN  & 0.16 & 84 \\

Strong form PINN with oversampling\cite{leung_nh-pinn_2022}  & $\approx$ 0.07 & -- \\

Strong form PINN\cite{leung_nh-pinn_2022}  & $\approx$ 0.3 & -- \\
\hline
\end{tabular}
\caption{The NRMSE and the training time of different models for solving the variable coefficient Poisson equation with periodic boundary conditions.}
\label{tab:Exp3}
\end{table}
\FloatBarrier
As shown in Table \ref{tab:Exp3}, the cell-based MLP with parameter sharing is almost two orders of magnitude more accurate compared to the strong form PINN with oversampling proposed in Leung et al. \cite{leung_nh-pinn_2022} and three orders of magnitude more accurate than standard strong form PINN.

\subsection{Poisson equation with high-frequency source term}
While solving the Poisson equation using the standard PINN, the PDE residual loss typically exhibits small magnitude high-frequency features that reduce numerical accuracy, whereas the boundary condition loss remains low. Wang et al. \cite{wang_multi-stage_2024} demonstrated that multi-stage neural networks can be used to improve accuracy; however, it requires the use of appropriate weight factors in the loss function and suitable scale factors for the MLP parameters. While training multi-stage networks, it is necessary to solve a Poisson equation with a high-frequency source function and zero boundary conditions. Therefore, in this example, we demonstrate that our cell-based MLP can resolve the Poisson equation with a high-frequency source function more effectively than the standard PINN, without the need to select a weight factor or scale factor. The corresponding strong form of the problem is:
\begin{align}
\begin{split}\label{eq:high_frequency_source_strong}
    \nabla u &= f(x,y),\\
    f(x,y) &= -\sin(6\pi x)\sin(6\pi y),
\end{split}    
\end{align}
where $(x,y)\in \Omega = [0,1]^2$, and the associated boundary conditions are
\begin{equation}
    u(0,y) = u(1,y) = u(x,0) = u(x,1) = 0.
\end{equation}
The true solution is
\begin{equation}
\label{eq:Exp4_true}
    u(x,y) = \frac{1}{2(6\pi)^2} \sin(6\pi x)\sin(6\pi y).
\end{equation}
The variational energy form of Eq. \eqref{eq:high_frequency_source_strong} is
\begin{equation}
    E = \int_\Omega \left[\frac{1}{2}\left( \nabla u\right)^2 + fu \right] \, d\Omega
\end{equation}

\begin{figure}
    \centering
    \includegraphics[width=1.0\textwidth]{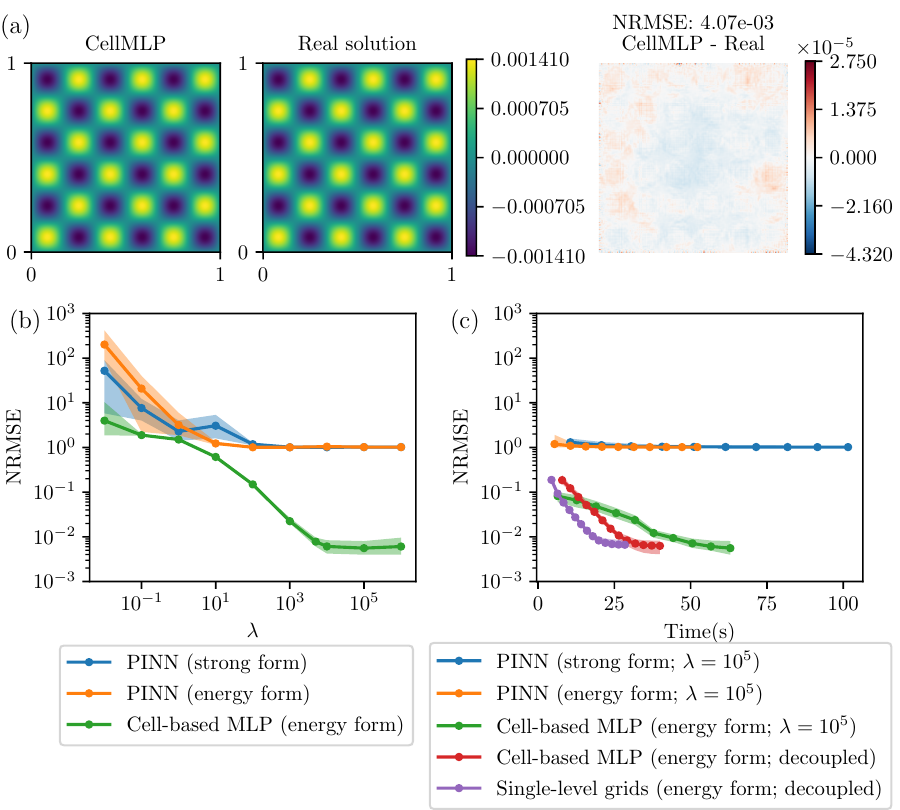}
    \caption{Model performance evaluation for the Poisson equation with high-frequency source term. (a) Solution of cell-based MLP with decoupled training scheme, analytical solution, and the difference between them; (b) The relative error (NRMSE) of the final solutions for different weight factor $\lambda$ in loss function for different models; (c) The relative error (NRMSE) during the training process for different models. }
    \label{fig:Exp4_results}
\end{figure}

As shown in Figure \ref{fig:Exp4_results}(a), the solution from the cell-based MLP with decoupled training scheme matches with the analytical solution (i.e., Eq. \eqref{eq:Exp4_true}), with the NRMSE around 0.4 \%. Similar to the results in Section \ref{sec:Exp1} and as discussed by Wang et al.\cite{wang_multi-stage_2024}, Figure \ref{fig:Exp4_results}(b) shows the weight factor $\lambda$ will significantly affect the solution error when enforcing the Dirichlet boundary condition by adding a penalty term in the loss function. Specifically, a $\lambda$ value of 1 results in an NRMSE exceeding 100 \%, whereas a $\lambda$ of $10^5$ lowers the NRMSE to about 0.4 \%. Even for the optimal value of $\lambda\approx 10^5$, the final error of the standard (strong/energy form) PINN is about 100\%. This is much greater than the cell-based MLP, which clearly demonstrates that the proposed model is a relatively better representation. As shown in Figure \ref{fig:Exp1_results}(c), if applying the decoupled training scheme, the cell-based MLP and the single-level grid converge faster and achieve higher accuracy compared to the strong and weak form PINNs, wherein the high-resolution grid is efficient for capturing the high-frequency features in the PDEs solution.  Additionally, if we choose the optimal value of $\lambda$, the cell-based MLP model without the decoupled training scheme will achieve a similar level of accuracy as the model with the decoupled training scheme, but it takes more computing time.

The multi-stage training scheme of Wang et al. \cite{wang_multi-stage_2024} includes algorithms for determining appropriate weight factors of loss functions and estimating a model scale factor and magnitude prefactor for linear PDEs with high-frequency source functions and zero boundary conditions. These algorithms allowed them to solve Eq. \eqref{eq:high_frequency_source_strong} with high accuracy, achieving a relative error of approximately 0.01 \%. 
Our study demonstrates that the cell-based MLPs can achieve a relative error of approximately 0.4 \% without needing to select weight factors of loss functions and scaling factors, but it is still significantly more accurate than the standard PINN that has a relative error of 100 \%. Because cell-based MLPs can be efficiently implemented using tiny-cuda-nn library (i.e., shorter compute times) than standard PINNs and multi-stage networks, they can be a promising alternative to solve multi-scale PDEs. 

\subsection{Hyperparameter}
Thus far, we have presented numerical examples demonstrating the advantages of using multiresolution cell-based MLP to solve Poisson problems with high-frequency terms and nonlinear/periodic boundary conditions. To establish the robustness of the proposed method and to guide the selection of hyperparameters, we now analyze how the hyperparameters influence the solution accuracy for the multiscale problem in Section \ref{sec:Exp2} with the decoupled training scheme and the periodic PDE in Section \ref{sec:Exp3} with the parameter-sharing scheme.

\subsubsection{Hyperparameters of the multiresolution grid cells}

\begin{figure}
    \centering
    \includegraphics[width=1\textwidth]{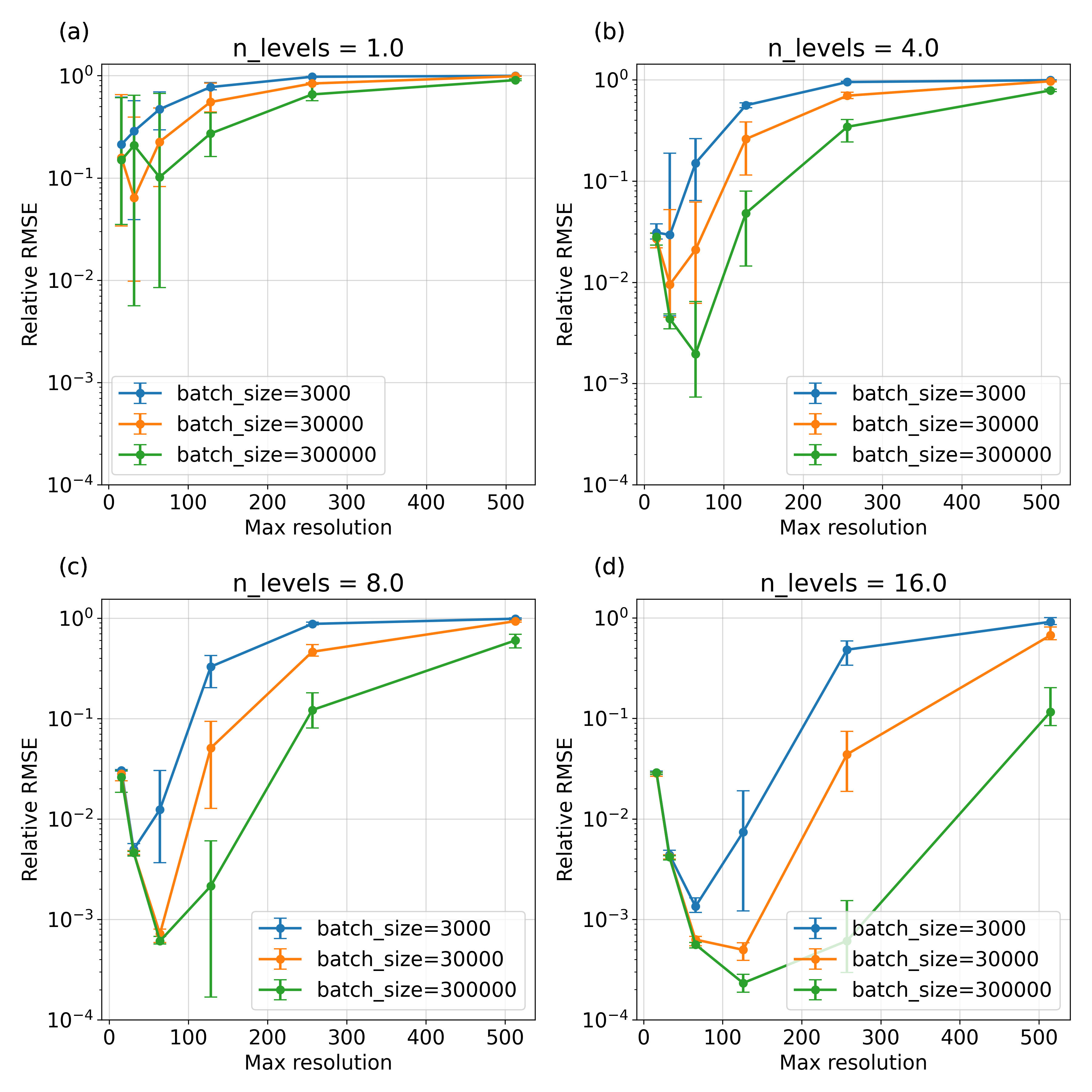}
    \caption{Influence of hyperparameters on solution accuracy with the number of grid levels (a) $L$ = 1; (b) $L$ = 4; (c) $L$ = 8; (d) $L$ = 16. The relative error of solving the multiscale PDE in Section \ref{sec:Exp2} is plotted for varying maximum cell resolution and different batch sizes of the training sample points.}
    \label{fig:Grid_hyperparameters_Exp2}
\end{figure}

\begin{figure}
    \centering
    \includegraphics[width=1\textwidth]{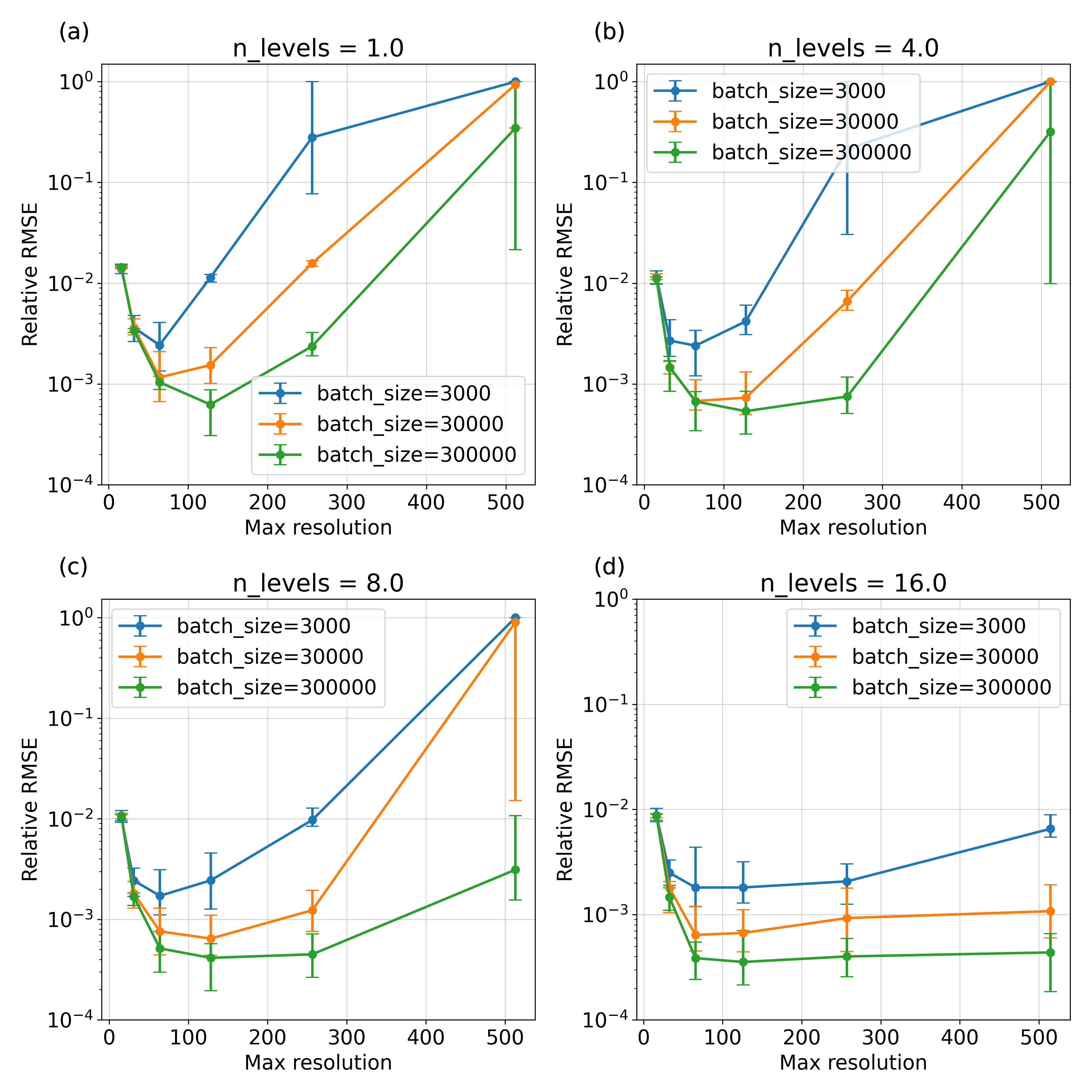}
    \caption{Influence of hyperparameters on solution accuracy with the number of grid levels (a) $L$ = 1; (b) $L$ = 4; (c) $L$ = 8; (d) $L$ = 16. The relative error of solving the periodic PDE in section \ref{sec:Exp3} is plotted for varying maximum cell resolution and different batch sizes of the training sample points.}
    \label{fig:Grid_hyperparameters_Exp3}
\end{figure}

We begin by investigating the influence of hyperparameters governing the grid cells by solving PDEs with hyperparameter ranges specified in Table \ref{tab:Grid_hyperparameters}. The influence of the hyperparameters of the multiresolution grid cells on the multiscale problem and the periodic PDE is illustrated in Figures \ref{fig:Grid_hyperparameters_Exp2} and \ref{fig:Grid_hyperparameters_Exp3}, respectively. 
From these studies, we generally observe that for a given number of grid levels and training batch size, if the maximum resolution is increased then the error in the solution gradually decreases until an optimal maximum resolution, beyond which the error gradually increases. However, if the training batch size is increased, both the optimal maximum resolution and solution accuracy increase correspondingly. This occurs because the grid cell needs to be exposed to a sufficient number of random sample points repeatedly for effective training. With a fixed training batch size, as the cell resolution is increased, the training batch size initially suffices to train the cells. Consequently, the finer grids enhance the model's representational capability, aiding in achieving a more accurate solution. Yet, beyond a certain resolution, the number of cells grows so large that some cells do not receive enough sample points for training, which leads to an increase in solution error as these undertrained cells negatively impact the accuracy. By increasing the training batch size, not only is the numerical integration for the variational energy more accurate, but more cells also receive adequate training. Thus, increasing the batch size along with the maximum resolution increases can help  decrease the solution error.

The provided results also demonstrated that an increase in the number of grid levels can lead to a reduction in errors for models that have a maximum resolution exceeding the optimal resolution. It is important to note, however, that the accuracy of the results varies depending on whether the model is solving a multiscale PDE with the decoupled training scheme or a periodic PDE with the parameter-sharing scheme. When solving the multiscale PDE with the decoupled training scheme, as depicted in Figure \ref{fig:Grid_hyperparameters_Exp2}, increasing the number of grid levels can lead to a significant reduction in the solution error with the optimal maximum resolution, as well as an increase in the optimal resolution. For example, the error decreases by two orders of magnitude when the number of levels $L$ is increased from 1 to 16 and the optimal resolution increases from a value less than 100 to greater than 100. Conversely, when solving the periodic PDE with the parameter-sharing scheme, as illustrated in Figure \ref{fig:Grid_hyperparameters_Exp3}, a higher number of grid levels has less impact on either the optimal resolution or the solution error of the model with the optimal resolution but can improve the solution accuracy for larger maximum resolutions.

The differences in the results presented in Figures \ref{fig:Grid_hyperparameters_Exp2} and \ref{fig:Grid_hyperparameters_Exp3} can be rationalized as follows. Increasing the number of grid levels in the model architecture with a fixed maximum resolution results in the use of more low-resolution grids in the multi-level approach. Similarly to the MLP, low-resolution grids can help the model better capture low-frequency features, enhancing the solution's smoothness. When using the decoupled training scheme to solve the multiscale PDE with fixed MLP parameters in the second training phase, the accuracy of the solution depends primarily on the representational capabilities of the multiresolution grids. Therefore, increasing the number of grid levels significantly improves the accuracy of the solution. In contrast, when solving the periodic PDE with the parameter-sharing scheme, the MLP parameters can also be trained to represent the smooth global component in the solution to some extent. Therefore, the addition of more low-resolution grids only slightly improves the results obtained by a relatively low-resolution model. In summary, using a larger batch size and an optimal maximum grid resolution with more multiresolution grids (i.e., multiple levels) enhances the model's performance.

\begin{table}[h!] 
\centering
\begin{tabular}{ >{\raggedright\arraybackslash}p{7cm}  >{\raggedright\arraybackslash}p{5cm}} 
\hline
Hyperparameter & Value \\
\hline
Number of grid levels ($L$) & 1,4,8,16  \\
Max cell resolution ($r_\text{max}$)  & 16,32,64,128,256,512  \\
Resolution ratio of adjacent levels ($\beta$)   & 1.12\\
Number of features per level ($F$) & 2\\
Batch size of training samples & $3\times10^{3}, 3\times10^{4}, 3\times10^{5}$ \\
Number of MLP hidden layers & 2 \\
Number of MLP neurons per layer & 32\\
\hline
\end{tabular}
\caption{Hyperparameters of the multiresolution cell-based MLP along with their ranges}
\label{tab:Grid_hyperparameters}
\end{table}
\FloatBarrier

\FloatBarrier

\subsubsection{Hyperparameters of the MLP and multiresolution grid cells}

We next investigated the impact of the hyperparameters of both the MLP and the multiresolution grid cells on solution accuracy. This was accomplished by exploring the hyperparameter ranges outlined in Table \ref{tab:Grid_MLP_hyperparameters}. 

\begin{figure}
    \centering
    \includegraphics[width=1\textwidth]{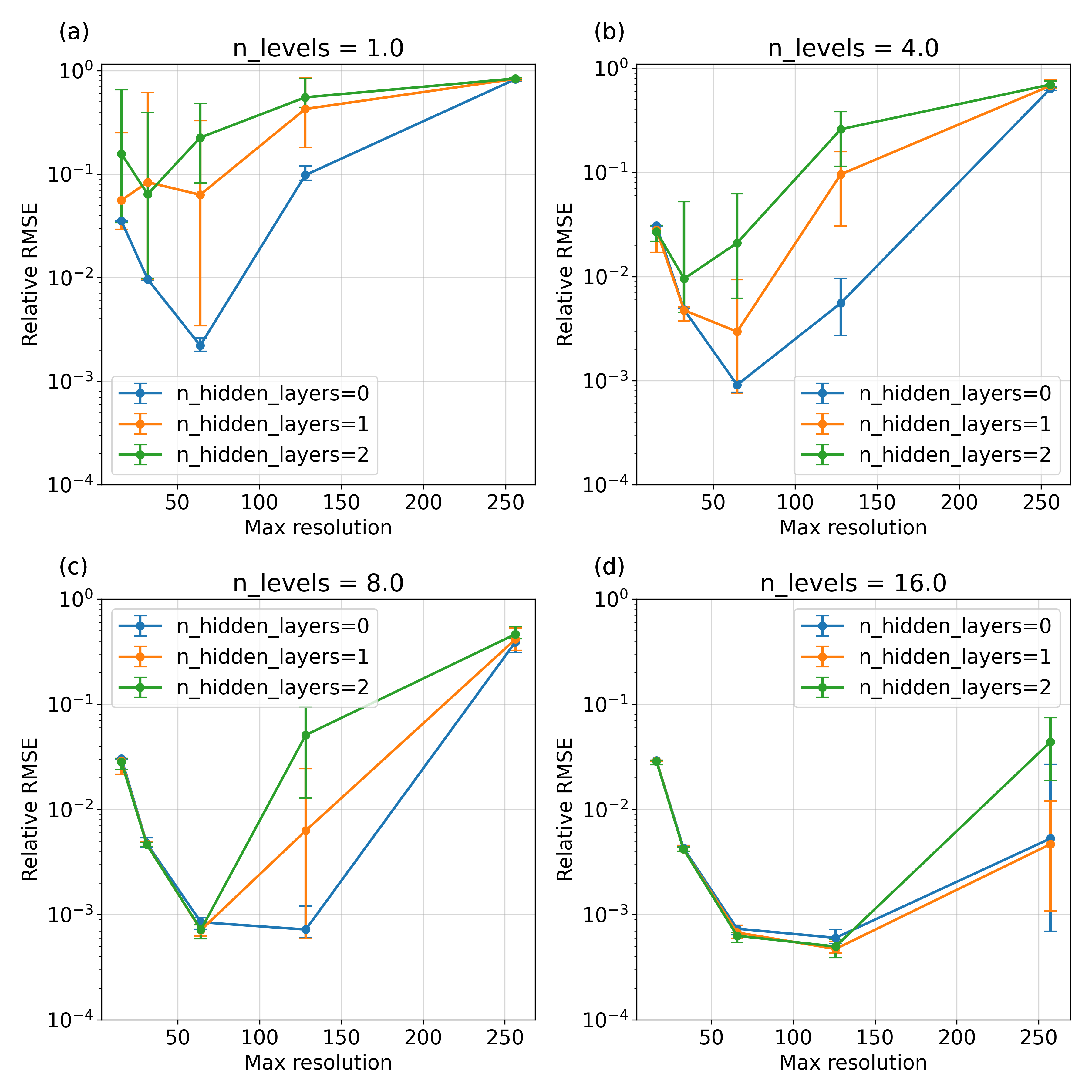}
    \caption{Influence of hyperparameters on solution accuracy with the number of grid levels (a) $L$ = 1; (b) $L$ = 4; (c) $L$ = 8; (d) $L$ = 16. The relative error of solving the multiscale PDE in Section \ref{sec:Exp2} is plotted for varying maximum cell resolution and number of MLP layers.}
    \label{fig:Grid_MLP_hyperparameters_Exp2}
\end{figure}

\begin{table}[h!] 
\centering
\begin{tabular}{ >{\raggedright\arraybackslash}p{7cm}  >{\raggedright\arraybackslash}p{5cm}} 
\hline
Hyperparameter & Value \\
\hline
Number of grid levels ($L$) & 1,4,8,16  \\
Max cell resolution ($r_\text{max}$)  & 16,32,64,128,256,512  \\
Resolution ratio of adjacent levels ($\beta$)   & 1.12\\
Number of features per level ($F$) & 2\\
Batch size of training samples & $3\times10^{4}$ \\
Number of MLP hidden layers & 0,1,2 \\
Number of MLP neurons per layer & 32\\
\hline
\end{tabular}
\caption{Hyperparameters of the multiresolution cell-based MLP along with their ranges}
\label{tab:Grid_MLP_hyperparameters}
\end{table}
\FloatBarrier
Figure \ref{fig:Grid_MLP_hyperparameters_Exp2} shows the results from solving the multiscale PDE using the decoupled training scheme. When the maximum resolution and the number of grid levels are held constant, a reduction in the number of MLP layers leads to a decrease in the model's solution error. However, increasing the number of grid levels tends to mitigate the discrepancies in results obtained with different numbers of MLP layers. This happens because the MLP helps in interpolating the solution field, similar to the shape functions in finite element methods. However, recall that in the decoupled training scheme, the MLP parameters are only trained to conform to the boundary conditions during the first phase of and remain fixed during the second phase. As a result, they do not adjust to better interpolate the solution field throughout the training process. Therefore, it is easier for the multiresolution grid to find the solution in the feature space with a simpler MLP (i.e., with fewer hidden layers). When the multiresolution grid is adequately large and has a sufficient number of grid levels, it is capable of accurately finding the solution in the feature space, even when integrated with a more complex interpolation function embodied by the MLP. From a computational standpoint, it can be beneficial to use 
a single hidden layer network integrated with the multiresolution grid while implementing the decoupled training scheme, depending upon the nonlinearity of the boundary condition. 

\begin{figure}
    \centering
    \includegraphics[width=1\textwidth]{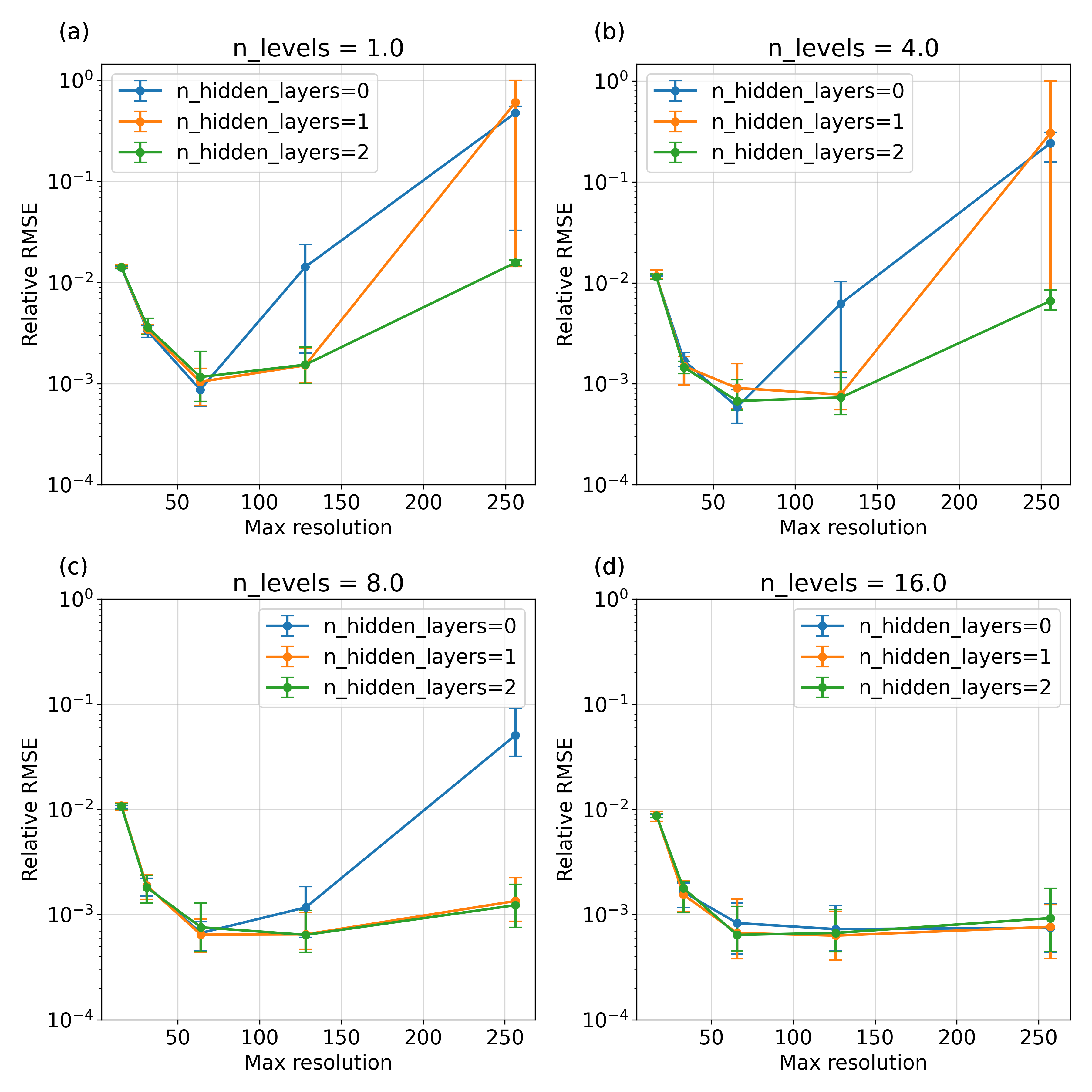}
    \caption{Influence of hyperparameters on solution accuracy with the number of grid levels (a) $L$ = 1; (b) $L$ = 4; (c) $L$ = 8; (d) $L$ = 16. The relative error of solving the periodic PDE in section \ref{sec:Exp3} is plotted for varying maximum cell resolution and number of MLP layers.}
    \label{fig:Grid_MLP_hyperparameters_Exp3}
\end{figure}

In contrast, when solving the periodic PDE with the parameter sharing scheme, as shown in Figure \ref{fig:Grid_MLP_hyperparameters_Exp3}, the MLP with more hidden layers can help reduce the solution error at larger maximum resolutions, thus improving model performance. Besides, increasing the number of grid levels tends to reduce the difference in results obtained with different numbers of MLP layers, which is similar to the trend in Figure \ref{fig:Grid_MLP_hyperparameters_Exp2}. As we mentioned before, this occurs because the MLP can help the model capture low-frequency features and improve the smoothness of the solution. Unlike in the decoupled training scheme, the MLP parameters are better trained in the parameter-sharing scheme, so a deeper MLP can improve the solution accuracy by helping the model capture the low-frequency features, especially if we use a smaller number of grid levels with a larger maximum cell resolution. In summary, this investigation suggests that taking a single hidden layer network is a better choice because, in general, we do not know the optimal maximum resolution \emph{a priori}, except when using a decoupled training scheme.

\section{Discussion}
In all the preceding studies, we restricted our attention to solving forward problems involving multiscale PDEs with bilinear interpolated multiresolution grids with MLPs on a square domain. However, PINNs and physics-informed cell representations may not be particularly advantageous for forward problems due to the higher computational cost. 
Instead, these physics-informed DL approaches show great potential for solving inverse problems compared to traditional numerical methods, such as the FEM. To make use of this potential with the cell-based MLP, we can integrate the data loss from laboratory experiments or field observations by leveraging the model architecture. One way to do this is to concatenate the inverse design parameter with the feature vector as the input to the MLPs. This idea was inspired by the approach used to solve Neural Radiance and Density Fields (NeRF) problems with a grid-based model \cite{muller_instant_2022}, where the material parameters were concatenated with the feature vector. 

Typically, in computational physics applications, especially in the various disciplines of earth and material sciences, PDEs need to be solved on irregularly shaped domains. For such problems, the current rectangular multiresolution grids are not well suited, so replacing the bilinear interpolation in rectangle cells with barycentric interpolation in triangles would be beneficial. For instance, if we intend to solve Stokes equations on arbitrarily shaped glaciers \cite{jouvet_inversion_2023, jimenez_updated-lagrangian_2017} or elasticity equations in heterogeneous material (e.g., fiber-reinforced composites \cite{zhang_analyses_2022}), the multiresolution grids in the background will not conform with the glacier or material domain shapes. This will result in large errors on the interface because the material properties vary sharply, often leading to steep gradients in the solution variables. However, by applying barycentric interpolation in triangles, we can create an appropriate multiresolution triangle mesh on the background that conforms with the circular interface. This can reduce the solution error and improve the overall performance of the model.

\section{Conclusion}
In this paper, we introduced a novel approach for solving multiscale PDEs through the use of physics-informed cell representations, whose architecture integrates multilevel multiresolution grids with MLP. We enhanced the efficiency by solving the variational form of the PDEs instead of the strong form, which allowed bilinear interpolation of feature vectors and effectively simplified the back propagation step involving derivative calculations. Additionally, we proposed a decoupled training scheme for enforcing Dirichlet boundary conditions and a parameter-sharing scheme for applying periodic boundary conditions, which effectively mitigated the issues posed by soft constraint-induced boundary (i.e., penalty) loss terms. Our empirical investigations reveal that this physics-informed cell representation or cell-based MLP model outperforms standard PINNs in terms of speed and accuracy. Especially for PDEs with multiscale features, we showed that standard PINNs are afflicted by slow convergence and poor accuracy, unlike cell-based MLPs. 

Our results demonstrate that cell representations can offer superior speed and accuracy for PDEs with nonlinear boundary conditions or high-frequency solutions. The decoupled training scheme allows the model to utilize the variational form of the PDE with nonlinear Dirichlet boundary conditions without the need for loss balancing. The parameter-sharing scheme allows the model to automatically satisfy the periodic boundary conditions, thereby substantially reducing the complexity of the optimization problem. Thus, the decoupled training method significantly expedites the learning process, whereas the parameter-sharing mechanism not only increases training speed but also diminishes solution errors in comparison to standard loss functions. We conducted comprehensive hyperparameter studies focusing on the multiresolution grid and MLPs within the frameworks of either a decoupled training or a parameter-sharing scheme. The findings from these experiments advocate for the use of larger batch sizes, a larger number of multilevel grids with an optimal maximum grid resolution, and the implementation of a shallow MLP (e.g., with a single hidden layer) for enhanced performance.

\section*{Data Availability}
The code required to reproduce the above findings is available to download from \href{https://github.com/YuxiangGao0321/Cell-based-MLP}{https://github.com/YuxiangGao0321/Cell-based-MLP}.

\section*{Acknowledgements}

This work was supported by a research grant from 3M Company. We would like to express our sincere gratitude to Dr. Matthew Berger for his invaluable discussions and insightful suggestions, which greatly contributed to the development of this work.



 \bibliographystyle{elsarticle-num}




\clearpage
\bibliography{manuscript}

\begin{thebibliography}{10}
\expandafter\ifx\csname url\endcsname\relax
  \def\url#1{\texttt{#1}}\fi
\expandafter\ifx\csname urlprefix\endcsname\relax\def\urlprefix{URL }\fi
\expandafter\ifx\csname href\endcsname\relax
  \def\href#1#2{#2} \def\path#1{#1}\fi

\bibitem{xu_multiscale_2021}
F.~Xu, H.~Hajibeygi, L.~J. Sluys, \href{https://linkinghub.elsevier.com/retrieve/pii/S0021999121001820}{Multiscale extended finite element method for deformable fractured porous media}, Journal of Computational Physics 436 (2021) 110287.
\newblock \href {https://doi.org/10.1016/j.jcp.2021.110287} {\path{doi:10.1016/j.jcp.2021.110287}}.
\newline\urlprefix\url{https://linkinghub.elsevier.com/retrieve/pii/S0021999121001820}

\bibitem{qian_multiscale_2004}
D.~Qian, G.~J. Wagner, W.~K. Liu, \href{https://linkinghub.elsevier.com/retrieve/pii/S0045782504000180}{A multiscale projection method for the analysis of carbon nanotubes}, Computer Methods in Applied Mechanics and Engineering 193~(17-20) (2004) 1603--1632.
\newblock \href {https://doi.org/10.1016/j.cma.2003.12.016} {\path{doi:10.1016/j.cma.2003.12.016}}.
\newline\urlprefix\url{https://linkinghub.elsevier.com/retrieve/pii/S0045782504000180}

\bibitem{abbaszadeh_reduced-order_2021}
M.~Abbaszadeh, M.~Dehghan, A.~Khodadadian, N.~Noii, C.~Heitzinger, T.~Wick, \href{https://linkinghub.elsevier.com/retrieve/pii/S0021999120306495}{A reduced-order variational multiscale interpolating element free {Galerkin} technique based on proper orthogonal decomposition for solving {Navier}–{Stokes} equations coupled with a heat transfer equation: {Nonstationary} incompressible {Boussinesq} equations}, Journal of Computational Physics 426 (2021) 109875.
\newblock \href {https://doi.org/10.1016/j.jcp.2020.109875} {\path{doi:10.1016/j.jcp.2020.109875}}.
\newline\urlprefix\url{https://linkinghub.elsevier.com/retrieve/pii/S0021999120306495}

\bibitem{hoang_conservative_2019}
T.-T.-P. Hoang, W.~Leng, L.~Ju, Z.~Wang, K.~Pieper, \href{https://linkinghub.elsevier.com/retrieve/pii/S0021999119300233}{Conservative explicit local time-stepping schemes for the shallow water equations}, Journal of Computational Physics 382 (2019) 152--176.
\newblock \href {https://doi.org/10.1016/j.jcp.2019.01.006} {\path{doi:10.1016/j.jcp.2019.01.006}}.
\newline\urlprefix\url{https://linkinghub.elsevier.com/retrieve/pii/S0021999119300233}

\bibitem{kim_direct_2024}
S.~Kim, K.~Saurabh, M.~A. Khanwale, A.~Mani, R.~K. Anand, B.~Ganapathysubramanian, \href{https://linkinghub.elsevier.com/retrieve/pii/S0021999123008434}{Direct numerical simulation of electrokinetic transport phenomena in fluids: {Variational} multi-scale stabilization and octree-based mesh refinement}, Journal of Computational Physics 500 (2024) 112747.
\newblock \href {https://doi.org/10.1016/j.jcp.2023.112747} {\path{doi:10.1016/j.jcp.2023.112747}}.
\newline\urlprefix\url{https://linkinghub.elsevier.com/retrieve/pii/S0021999123008434}

\bibitem{yu_phase_2022}
Q.~Yu, Q.~Xia, Y.~Li, \href{https://linkinghub.elsevier.com/retrieve/pii/S0021999122004454}{A phase field-based systematic multiscale topology optimization method for porous structures design}, Journal of Computational Physics 466 (2022) 111383.
\newblock \href {https://doi.org/10.1016/j.jcp.2022.111383} {\path{doi:10.1016/j.jcp.2022.111383}}.
\newline\urlprefix\url{https://linkinghub.elsevier.com/retrieve/pii/S0021999122004454}

\bibitem{xu_multiscale_2016}
F.~Xu, M.~Gunzburger, J.~Burkardt, Q.~Du, \href{http://epubs.siam.org/doi/10.1137/15M1010300}{A {Multiscale} {Implementation} {Based} on {Adaptive} {Mesh} {Refinement} for the {Nonlocal} {Peridynamics} {Model} in {One} {Dimension}}, Multiscale Modeling \& Simulation 14~(1) (2016) 398--429.
\newblock \href {https://doi.org/10.1137/15M1010300} {\path{doi:10.1137/15M1010300}}.
\newline\urlprefix\url{http://epubs.siam.org/doi/10.1137/15M1010300}

\bibitem{li_second-order_2016}
Z.-H. Li, Q.~Ma, J.~Cui, \href{https://linkinghub.elsevier.com/retrieve/pii/S0021999116001868}{Second-order two-scale finite element algorithm for dynamic thermo–mechanical coupling problem in symmetric structure}, Journal of Computational Physics 314 (2016) 712--748.
\newblock \href {https://doi.org/10.1016/j.jcp.2016.03.034} {\path{doi:10.1016/j.jcp.2016.03.034}}.
\newline\urlprefix\url{https://linkinghub.elsevier.com/retrieve/pii/S0021999116001868}

\bibitem{chen_high_2023}
W.~Chen, K.~Wu, T.~Xiong, \href{https://linkinghub.elsevier.com/retrieve/pii/S0021999123003352}{High order asymptotic preserving finite difference {WENO} schemes with constrained transport for {MHD} equations in all sonic {Mach} numbers}, Journal of Computational Physics 488 (2023) 112240.
\newblock \href {https://doi.org/10.1016/j.jcp.2023.112240} {\path{doi:10.1016/j.jcp.2023.112240}}.
\newline\urlprefix\url{https://linkinghub.elsevier.com/retrieve/pii/S0021999123003352}

\bibitem{asad_mechanics-informed_2023}
F.~As’ad, C.~Farhat, \href{https://linkinghub.elsevier.com/retrieve/pii/S004578252300587X}{A mechanics-informed deep learning framework for data-driven nonlinear viscoelasticity}, Computer Methods in Applied Mechanics and Engineering 417 (2023) 116463.
\newblock \href {https://doi.org/10.1016/j.cma.2023.116463} {\path{doi:10.1016/j.cma.2023.116463}}.
\newline\urlprefix\url{https://linkinghub.elsevier.com/retrieve/pii/S004578252300587X}

\bibitem{taneja_multi-resolution_2023}
K.~Taneja, X.~He, Q.~He, J.-S. Chen, \href{https://link.springer.com/10.1007/s00466-023-02403-x}{A multi-resolution physics-informed recurrent neural network: formulation and application to musculoskeletal systems}, Computational Mechanics (Oct. 2023).
\newblock \href {https://doi.org/10.1007/s00466-023-02403-x} {\path{doi:10.1007/s00466-023-02403-x}}.
\newline\urlprefix\url{https://link.springer.com/10.1007/s00466-023-02403-x}

\bibitem{park_convolution_2023}
C.~Park, Y.~Lu, S.~Saha, T.~Xue, J.~Guo, S.~Mojumder, D.~W. Apley, G.~J. Wagner, W.~K. Liu, \href{https://link.springer.com/10.1007/s00466-023-02329-4}{Convolution hierarchical deep-learning neural network ({C}-{HiDeNN}) with graphics processing unit ({GPU}) acceleration}, Computational Mechanics 72~(2) (2023) 383--409.
\newblock \href {https://doi.org/10.1007/s00466-023-02329-4} {\path{doi:10.1007/s00466-023-02329-4}}.
\newline\urlprefix\url{https://link.springer.com/10.1007/s00466-023-02329-4}

\bibitem{xiao_geometric_2023}
M.~Xiao, R.~Ma, W.~Sun, \href{https://linkinghub.elsevier.com/retrieve/pii/S0045782523003432}{Geometric learning for computational mechanics, {Part} {III}: {Physics}-constrained response surface of geometrically nonlinear shells}, Computer Methods in Applied Mechanics and Engineering 415 (2023) 116219.
\newblock \href {https://doi.org/10.1016/j.cma.2023.116219} {\path{doi:10.1016/j.cma.2023.116219}}.
\newline\urlprefix\url{https://linkinghub.elsevier.com/retrieve/pii/S0045782523003432}

\bibitem{liu_hidenn-fem_2023}
Y.~Liu, C.~Park, Y.~Lu, S.~Mojumder, W.~K. Liu, D.~Qian, \href{https://link.springer.com/10.1007/s00466-023-02293-z}{{HiDeNN}-{FEM}: a seamless machine learning approach to nonlinear finite element analysis}, Computational Mechanics 72~(1) (2023) 173--194.
\newblock \href {https://doi.org/10.1007/s00466-023-02293-z} {\path{doi:10.1007/s00466-023-02293-z}}.
\newline\urlprefix\url{https://link.springer.com/10.1007/s00466-023-02293-z}

\bibitem{xue_physics-embedded_2022}
T.~Xue, Z.~Gan, S.~Liao, J.~Cao, \href{https://www.nature.com/articles/s41524-022-00890-9}{Physics-embedded graph network for accelerating phase-field simulation of microstructure evolution in additive manufacturing}, npj Computational Materials 8~(1) (2022) 201.
\newblock \href {https://doi.org/10.1038/s41524-022-00890-9} {\path{doi:10.1038/s41524-022-00890-9}}.
\newline\urlprefix\url{https://www.nature.com/articles/s41524-022-00890-9}

\bibitem{hernandez_port-metriplectic_2023}
Q.~Hernández, A.~Badías, F.~Chinesta, E.~Cueto, \href{https://link.springer.com/10.1007/s00466-023-02296-w}{Port-metriplectic neural networks: thermodynamics-informed machine learning of complex physical systems}, Computational Mechanics 72~(3) (2023) 553--561.
\newblock \href {https://doi.org/10.1007/s00466-023-02296-w} {\path{doi:10.1007/s00466-023-02296-w}}.
\newline\urlprefix\url{https://link.springer.com/10.1007/s00466-023-02296-w}

\bibitem{yang_using_2023}
M.~Yang, J.~T. Foster, \href{https://www.dl.begellhouse.com/journals/558048804a15188a,01cca90e1b73210a,04cad08313f652a7.html}{{USING} {PHYSICS}-{INFORMED} {NEURAL} {NETWORKS} {TO} {SOLVE} {FOR} {PERMEABILITY} {FIELD} {UNDER} {TWO}-{PHASE} {FLOW} {IN} {HETEROGENEOUS} {POROUS} {MEDIA}}, Journal of Machine Learning for Modeling and Computing 4~(1) (2023) 1--19.
\newblock \href {https://doi.org/10.1615/JMachLearnModelComput.2023046921} {\path{doi:10.1615/JMachLearnModelComput.2023046921}}.
\newline\urlprefix\url{https://www.dl.begellhouse.com/journals/558048804a15188a,01cca90e1b73210a,04cad08313f652a7.html}

\bibitem{gao_cnn-based_2023}
Y.~Gao, M.~Berger, R.~Duddu, \href{https://ascelibrary.org/doi/10.1061/JENMDT.EMENG-6936}{{CNN}-{Based} {Surrogate} for the {Phase} {Field} {Damage} {Model}: {Generalization} across {Microstructure} {Parameters} for {Composite} {Materials}}, Journal of Engineering Mechanics 149~(6) (2023) 04023025.
\newblock \href {https://doi.org/10.1061/JENMDT.EMENG-6936} {\path{doi:10.1061/JENMDT.EMENG-6936}}.
\newline\urlprefix\url{https://ascelibrary.org/doi/10.1061/JENMDT.EMENG-6936}

\bibitem{azizzadenesheli_neural_2024}
K.~Azizzadenesheli, N.~Kovachki, Z.~Li, M.~Liu-Schiaffini, J.~Kossaifi, A.~Anandkumar, \href{https://www.nature.com/articles/s42254-024-00712-5}{Neural operators for accelerating scientific simulations and design}, Nature Reviews Physics (2024) 1--9\href {https://doi.org/10.1038/s42254-024-00712-5} {\path{doi:10.1038/s42254-024-00712-5}}.
\newline\urlprefix\url{https://www.nature.com/articles/s42254-024-00712-5}

\bibitem{geng_deep_2024}
Y.~Geng, Y.~Teng, Z.~Wang, L.~Ju, \href{https://linkinghub.elsevier.com/retrieve/pii/S0021999123006848}{A deep learning method for the dynamics of classic and conservative {Allen}-{Cahn} equations based on fully-discrete operators}, Journal of Computational Physics 496 (2024) 112589.
\newblock \href {https://doi.org/10.1016/j.jcp.2023.112589} {\path{doi:10.1016/j.jcp.2023.112589}}.
\newline\urlprefix\url{https://linkinghub.elsevier.com/retrieve/pii/S0021999123006848}

\bibitem{perera_graph_2022}
R.~Perera, D.~Guzzetti, V.~Agrawal, \href{https://linkinghub.elsevier.com/retrieve/pii/S004578252200250X}{Graph neural networks for simulating crack coalescence and propagation in brittle materials}, Computer Methods in Applied Mechanics and Engineering 395 (2022) 115021.
\newblock \href {https://doi.org/10.1016/j.cma.2022.115021} {\path{doi:10.1016/j.cma.2022.115021}}.
\newline\urlprefix\url{https://linkinghub.elsevier.com/retrieve/pii/S004578252200250X}

\bibitem{raissi_physics-informed_2019}
M.~Raissi, P.~Perdikaris, G.~Karniadakis, \href{https://linkinghub.elsevier.com/retrieve/pii/S0021999118307125}{Physics-informed neural networks: {A} deep learning framework for solving forward and inverse problems involving nonlinear partial differential equations}, Journal of Computational Physics 378 (2019) 686--707.
\newblock \href {https://doi.org/10.1016/j.jcp.2018.10.045} {\path{doi:10.1016/j.jcp.2018.10.045}}.
\newline\urlprefix\url{https://linkinghub.elsevier.com/retrieve/pii/S0021999118307125}

\bibitem{yuan_-pinn_2022}
L.~Yuan, Y.-Q. Ni, X.-Y. Deng, S.~Hao, \href{https://www.sciencedirect.com/science/article/pii/S0021999122003229}{A-{PINN}: {Auxiliary} physics informed neural networks for forward and inverse problems of nonlinear integro-differential equations}, Journal of Computational Physics 462 (2022) 111260.
\newblock \href {https://doi.org/10.1016/j.jcp.2022.111260} {\path{doi:10.1016/j.jcp.2022.111260}}.
\newline\urlprefix\url{https://www.sciencedirect.com/science/article/pii/S0021999122003229}

\bibitem{mowlavi_optimal_2023}
S.~Mowlavi, S.~Nabi, \href{https://www.sciencedirect.com/science/article/pii/S002199912200794X}{Optimal control of {PDEs} using physics-informed neural networks}, Journal of Computational Physics 473 (2023) 111731.
\newblock \href {https://doi.org/10.1016/j.jcp.2022.111731} {\path{doi:10.1016/j.jcp.2022.111731}}.
\newline\urlprefix\url{https://www.sciencedirect.com/science/article/pii/S002199912200794X}

\bibitem{gao_phygeonet_2021}
H.~Gao, L.~Sun, J.-X. Wang, \href{https://www.sciencedirect.com/science/article/pii/S0021999120308536}{{PhyGeoNet}: {Physics}-informed geometry-adaptive convolutional neural networks for solving parameterized steady-state {PDEs} on irregular domain}, Journal of Computational Physics 428 (2021) 110079.
\newblock \href {https://doi.org/10.1016/j.jcp.2020.110079} {\path{doi:10.1016/j.jcp.2020.110079}}.
\newline\urlprefix\url{https://www.sciencedirect.com/science/article/pii/S0021999120308536}

\bibitem{chen_physics-informed_2021}
W.~Chen, Q.~Wang, J.~S. Hesthaven, C.~Zhang, \href{https://www.sciencedirect.com/science/article/pii/S0021999121005611}{Physics-informed machine learning for reduced-order modeling of nonlinear problems}, Journal of Computational Physics 446 (2021) 110666.
\newblock \href {https://doi.org/10.1016/j.jcp.2021.110666} {\path{doi:10.1016/j.jcp.2021.110666}}.
\newline\urlprefix\url{https://www.sciencedirect.com/science/article/pii/S0021999121005611}

\bibitem{lucor_simple_2022}
D.~Lucor, A.~Agrawal, A.~Sergent, \href{https://www.sciencedirect.com/science/article/pii/S0021999122000845}{Simple computational strategies for more effective physics-informed neural networks modeling of turbulent natural convection}, Journal of Computational Physics 456 (2022) 111022.
\newblock \href {https://doi.org/10.1016/j.jcp.2022.111022} {\path{doi:10.1016/j.jcp.2022.111022}}.
\newline\urlprefix\url{https://www.sciencedirect.com/science/article/pii/S0021999122000845}

\bibitem{khara_neural_2024}
B.~Khara, E.~Herron, A.~Balu, D.~Gamdha, C.-H. Yang, K.~Saurabh, A.~Jignasu, Z.~Jiang, S.~Sarkar, C.~Hegde, B.~Ganapathysubramanian, A.~Krishnamurthy, \href{https://linkinghub.elsevier.com/retrieve/pii/S0010448524000368}{Neural {PDE} {Solvers} for {Irregular} {Domains}}, Computer-Aided Design 172 (2024) 103709.
\newblock \href {https://doi.org/10.1016/j.cad.2024.103709} {\path{doi:10.1016/j.cad.2024.103709}}.
\newline\urlprefix\url{https://linkinghub.elsevier.com/retrieve/pii/S0010448524000368}

\bibitem{yang_multi-output_2022}
M.~Yang, J.~T. Foster, \href{https://linkinghub.elsevier.com/retrieve/pii/S0045782522002602}{Multi-output physics-informed neural networks for forward and inverse {PDE} problems with uncertainties}, Computer Methods in Applied Mechanics and Engineering 402 (2022) 115041.
\newblock \href {https://doi.org/10.1016/j.cma.2022.115041} {\path{doi:10.1016/j.cma.2022.115041}}.
\newline\urlprefix\url{https://linkinghub.elsevier.com/retrieve/pii/S0045782522002602}

\bibitem{rahaman_spectral_2019}
N.~Rahaman, A.~Baratin, D.~Arpit, F.~Draxler, M.~Lin, F.~Hamprecht, Y.~Bengio, A.~Courville, \href{https://proceedings.mlr.press/v97/rahaman19a.html}{On the {Spectral} {Bias} of {Neural} {Networks}}, in: Proceedings of the 36th {International} {Conference} on {Machine} {Learning}, PMLR, 2019, pp. 5301--5310.
\newline\urlprefix\url{https://proceedings.mlr.press/v97/rahaman19a.html}

\bibitem{tancik2020fourier}
M.~Tancik, P.~Srinivasan, B.~Mildenhall, S.~Fridovich-Keil, N.~Raghavan, U.~Singhal, R.~Ramamoorthi, J.~Barron, R.~Ng, Fourier features let networks learn high frequency functions in low dimensional domains, Advances in neural information processing systems 33 (2020) 7537--7547.

\bibitem{moseley_finite_2023}
B.~Moseley, A.~Markham, T.~Nissen-Meyer, \href{https://link.springer.com/10.1007/s10444-023-10065-9}{Finite basis physics-informed neural networks ({FBPINNs}): a scalable domain decomposition approach for solving differential equations}, Advances in Computational Mathematics 49~(4) (2023) 62.
\newblock \href {https://doi.org/10.1007/s10444-023-10065-9} {\path{doi:10.1007/s10444-023-10065-9}}.
\newline\urlprefix\url{https://link.springer.com/10.1007/s10444-023-10065-9}

\bibitem{wang_eigenvector_2021}
S.~Wang, H.~Wang, P.~Perdikaris, \href{https://linkinghub.elsevier.com/retrieve/pii/S0045782521002759}{On the eigenvector bias of {Fourier} feature networks: {From} regression to solving multi-scale {PDEs} with physics-informed neural networks}, Computer Methods in Applied Mechanics and Engineering 384 (2021) 113938.
\newblock \href {https://doi.org/10.1016/j.cma.2021.113938} {\path{doi:10.1016/j.cma.2021.113938}}.
\newline\urlprefix\url{https://linkinghub.elsevier.com/retrieve/pii/S0045782521002759}

\bibitem{krishnapriyan_characterizing_2021}
A.~Krishnapriyan, A.~Gholami, S.~Zhe, R.~Kirby, M.~W. Mahoney, \href{https://proceedings.neurips.cc/paper/2021/hash/df438e5206f31600e6ae4af72f2725f1-Abstract.html}{Characterizing possible failure modes in physics-informed neural networks}, in: Advances in {Neural} {Information} {Processing} {Systems}, Vol.~34, Curran Associates, Inc., 2021, pp. 26548--26560.
\newline\urlprefix\url{https://proceedings.neurips.cc/paper/2021/hash/df438e5206f31600e6ae4af72f2725f1-Abstract.html}

\bibitem{leung_nh-pinn_2022}
W.~T. Leung, G.~Lin, Z.~Zhang, \href{https://linkinghub.elsevier.com/retrieve/pii/S0021999122006015}{{NH}-{PINN}: {Neural} homogenization-based physics-informed neural network for multiscale problems}, Journal of Computational Physics 470 (2022) 111539.
\newblock \href {https://doi.org/10.1016/j.jcp.2022.111539} {\path{doi:10.1016/j.jcp.2022.111539}}.
\newline\urlprefix\url{https://linkinghub.elsevier.com/retrieve/pii/S0021999122006015}

\bibitem{wang_multi-stage_2024}
Y.~Wang, C.-Y. Lai, \href{https://linkinghub.elsevier.com/retrieve/pii/S0021999124001141}{Multi-stage neural networks: {Function} approximator of machine precision}, Journal of Computational Physics 504 (2024) 112865.
\newblock \href {https://doi.org/10.1016/j.jcp.2024.112865} {\path{doi:10.1016/j.jcp.2024.112865}}.
\newline\urlprefix\url{https://linkinghub.elsevier.com/retrieve/pii/S0021999124001141}

\bibitem{wang_when_2022}
S.~Wang, X.~Yu, P.~Perdikaris, \href{https://linkinghub.elsevier.com/retrieve/pii/S002199912100663X}{When and why {PINNs} fail to train: {A} neural tangent kernel perspective}, Journal of Computational Physics 449 (2022) 110768.
\newblock \href {https://doi.org/10.1016/j.jcp.2021.110768} {\path{doi:10.1016/j.jcp.2021.110768}}.
\newline\urlprefix\url{https://linkinghub.elsevier.com/retrieve/pii/S002199912100663X}

\bibitem{goswami_transfer_2020}
S.~Goswami, C.~Anitescu, S.~Chakraborty, T.~Rabczuk, \href{https://linkinghub.elsevier.com/retrieve/pii/S016784421930357X}{Transfer learning enhanced physics informed neural network for phase-field modeling of fracture}, Theoretical and Applied Fracture Mechanics 106 (2020) 102447.
\newblock \href {https://doi.org/10.1016/j.tafmec.2019.102447} {\path{doi:10.1016/j.tafmec.2019.102447}}.
\newline\urlprefix\url{https://linkinghub.elsevier.com/retrieve/pii/S016784421930357X}

\bibitem{samaniego_energy_2020}
E.~Samaniego, C.~Anitescu, S.~Goswami, V.~M. Nguyen-Thanh, H.~Guo, K.~Hamdia, T.~Rabczuk, X.~Zhuang, \href{http://arxiv.org/abs/1908.10407}{An {Energy} {Approach} to the {Solution} of {Partial} {Differential} {Equations} in {Computational} {Mechanics} via {Machine} {Learning}: {Concepts}, {Implementation} and {Applications}}, Computer Methods in Applied Mechanics and Engineering 362 (2020) 112790.
\newblock \href {https://doi.org/10.1016/j.cma.2019.112790} {\path{doi:10.1016/j.cma.2019.112790}}.
\newline\urlprefix\url{http://arxiv.org/abs/1908.10407}

\bibitem{schiassi_extreme_2021}
E.~Schiassi, R.~Furfaro, C.~Leake, M.~De~Florio, H.~Johnston, D.~Mortari, \href{https://linkinghub.elsevier.com/retrieve/pii/S0925231221009140}{Extreme theory of functional connections: {A} fast physics-informed neural network method for solving ordinary and partial differential equations}, Neurocomputing 457 (2021) 334--356.
\newblock \href {https://doi.org/10.1016/j.neucom.2021.06.015} {\path{doi:10.1016/j.neucom.2021.06.015}}.
\newline\urlprefix\url{https://linkinghub.elsevier.com/retrieve/pii/S0925231221009140}

\bibitem{mortari_multivariate_2019}
D.~Mortari, C.~Leake, \href{https://www.mdpi.com/2227-7390/7/3/296}{The {Multivariate} {Theory} of {Connections}}, Mathematics 7~(3) (2019) 296.
\newblock \href {https://doi.org/10.3390/math7030296} {\path{doi:10.3390/math7030296}}.
\newline\urlprefix\url{https://www.mdpi.com/2227-7390/7/3/296}

\bibitem{xie_automatic_2023}
Y.~Xie, Y.~Ma, Y.~Wang, \href{https://linkinghub.elsevier.com/retrieve/pii/S0045782523002633}{Automatic boundary fitting framework of boundary dependent physics-informed neural network solving partial differential equation with complex boundary conditions}, Computer Methods in Applied Mechanics and Engineering 414 (2023) 116139.
\newblock \href {https://doi.org/10.1016/j.cma.2023.116139} {\path{doi:10.1016/j.cma.2023.116139}}.
\newline\urlprefix\url{https://linkinghub.elsevier.com/retrieve/pii/S0045782523002633}

\bibitem{takikawa_neural_2021}
T.~Takikawa, J.~Litalien, K.~Yin, K.~Kreis, C.~Loop, D.~Nowrouzezahrai, A.~Jacobson, M.~McGuire, S.~Fidler, \href{https://ieeexplore.ieee.org/document/9578205/}{Neural {Geometric} {Level} of {Detail}: {Real}-time {Rendering} with {Implicit} {3D} {Shapes}}, in: 2021 {IEEE}/{CVF} {Conference} on {Computer} {Vision} and {Pattern} {Recognition} ({CVPR}), IEEE, Nashville, TN, USA, 2021, pp. 11353--11362.
\newblock \href {https://doi.org/10.1109/CVPR46437.2021.01120} {\path{doi:10.1109/CVPR46437.2021.01120}}.
\newline\urlprefix\url{https://ieeexplore.ieee.org/document/9578205/}

\bibitem{hadadan_neural_2021}
S.~Hadadan, S.~Chen, M.~Zwicker, \href{https://dl.acm.org/doi/10.1145/3478513.3480569}{Neural radiosity}, ACM Transactions on Graphics 40~(6) (2021) 1--11.
\newblock \href {https://doi.org/10.1145/3478513.3480569} {\path{doi:10.1145/3478513.3480569}}.
\newline\urlprefix\url{https://dl.acm.org/doi/10.1145/3478513.3480569}

\bibitem{muller_instant_2022}
T.~Müller, A.~Evans, C.~Schied, A.~Keller, \href{https://dl.acm.org/doi/10.1145/3528223.3530127}{Instant neural graphics primitives with a multiresolution hash encoding}, ACM Transactions on Graphics 41~(4) (2022) 1--15.
\newblock \href {https://doi.org/10.1145/3528223.3530127} {\path{doi:10.1145/3528223.3530127}}.
\newline\urlprefix\url{https://dl.acm.org/doi/10.1145/3528223.3530127}

\bibitem{kang_pixel_2023}
N.~Kang, B.~Lee, Y.~Hong, S.-B. Yun, E.~Park, \href{http://arxiv.org/abs/2207.12800}{{PIXEL}: {Physics}-{Informed} {Cell} {Representations} for {Fast} and {Accurate} {PDE} {Solvers}} (Feb. 2023).
\newline\urlprefix\url{http://arxiv.org/abs/2207.12800}

\bibitem{e_deep_2018}
W.~E, B.~Yu, \href{http://link.springer.com/10.1007/s40304-018-0127-z}{The {Deep} {Ritz} {Method}: {A} {Deep} {Learning}-{Based} {Numerical} {Algorithm} for {Solving} {Variational} {Problems}}, Communications in Mathematics and Statistics 6~(1) (2018) 1--12.
\newblock \href {https://doi.org/10.1007/s40304-018-0127-z} {\path{doi:10.1007/s40304-018-0127-z}}.
\newline\urlprefix\url{http://link.springer.com/10.1007/s40304-018-0127-z}

\bibitem{dacorogna_introduction_2014}
B.~Dacorogna, \href{https://www.worldscientific.com/worldscibooks/10.1142/p967}{Introduction to the {Calculus} of {Variations}}, 3rd Edition, IMPERIAL COLLEGE PRESS, 2014.
\newline\urlprefix\url{https://www.worldscientific.com/worldscibooks/10.1142/p967}

\bibitem{matous_review_2017}
K.~Matouš, M.~G.~D. Geers, V.~G. Kouznetsova, A.~Gillman, \href{https://www.sciencedirect.com/science/article/pii/S0021999116305782}{A review of predictive nonlinear theories for multiscale modeling of heterogeneous materials}, Journal of Computational Physics 330 (2017) 192--220.
\newblock \href {https://doi.org/10.1016/j.jcp.2016.10.070} {\path{doi:10.1016/j.jcp.2016.10.070}}.
\newline\urlprefix\url{https://www.sciencedirect.com/science/article/pii/S0021999116305782}

\bibitem{miyato_spectral_2018}
T.~Miyato, T.~Kataoka, M.~Koyama, Y.~Yoshida, \href{http://arxiv.org/abs/1802.05957}{Spectral {Normalization} for {Generative} {Adversarial} {Networks}} (Feb. 2018).
\newline\urlprefix\url{http://arxiv.org/abs/1802.05957}

\bibitem{muller_tiny-cuda-nn_2021}
T.~Müller, \href{https://github.com/NVlabs/tiny-cuda-nn}{tiny-cuda-nn} (Apr. 2021).
\newline\urlprefix\url{https://github.com/NVlabs/tiny-cuda-nn}

\bibitem{kingma2014adam}
D.~P. Kingma, J.~Ba, Adam: A method for stochastic optimization, arXiv preprint arXiv:1412.6980 (2014).

\bibitem{alnaes_fenics_2015}
M.~Alnæs, J.~Blechta, J.~Hake, A.~Johansson, B.~Kehlet, A.~Logg, C.~Richardson, J.~Ring, M.~E. Rognes, G.~N. Wells, \href{https://journals.ub.uni-heidelberg.de/index.php/ans/article/view/20553}{The {FEniCS} {Project} {Version} 1.5}, Archive of Numerical Software 3~(100) (Dec. 2015).
\newblock \href {https://doi.org/10.11588/ans.2015.100.20553} {\path{doi:10.11588/ans.2015.100.20553}}.
\newline\urlprefix\url{https://journals.ub.uni-heidelberg.de/index.php/ans/article/view/20553}

\bibitem{lagaris_artificial_1998}
I.~E. Lagaris, A.~Likas, D.~I. Fotiadis, \href{http://arxiv.org/abs/physics/9705023}{Artificial {Neural} {Networks} for {Solving} {Ordinary} and {Partial} {Differential} {Equations}}, IEEE Transactions on Neural Networks 9~(5) (1998) 987--1000.
\newblock \href {https://doi.org/10.1109/72.712178} {\path{doi:10.1109/72.712178}}.
\newline\urlprefix\url{http://arxiv.org/abs/physics/9705023}

\bibitem{jouvet_inversion_2023}
G.~Jouvet, \href{https://www.cambridge.org/core/product/identifier/S0022143022000417/type/journal_article}{Inversion of a {Stokes} glacier flow model emulated by deep learning}, Journal of Glaciology 69~(273) (2023) 13--26.
\newblock \href {https://doi.org/10.1017/jog.2022.41} {\path{doi:10.1017/jog.2022.41}}.
\newline\urlprefix\url{https://www.cambridge.org/core/product/identifier/S0022143022000417/type/journal_article}

\bibitem{jimenez_updated-lagrangian_2017}
S.~Jiménez, R.~Duddu, J.~Bassis, \href{https://linkinghub.elsevier.com/retrieve/pii/S0045782516303140}{An updated-{Lagrangian} damage mechanics formulation for modeling the creeping flow and fracture of ice sheets}, Computer Methods in Applied Mechanics and Engineering 313 (2017) 406--432.
\newblock \href {https://doi.org/10.1016/j.cma.2016.09.034} {\path{doi:10.1016/j.cma.2016.09.034}}.
\newline\urlprefix\url{https://linkinghub.elsevier.com/retrieve/pii/S0045782516303140}

\bibitem{zhang_analyses_2022}
E.~Zhang, M.~Dao, G.~E. Karniadakis, S.~Suresh, \href{https://www.science.org/doi/10.1126/sciadv.abk0644}{Analyses of internal structures and defects in materials using physics-informed neural networks}, Science Advances 8~(7) (2022) eabk0644.
\newblock \href {https://doi.org/10.1126/sciadv.abk0644} {\path{doi:10.1126/sciadv.abk0644}}.
\newline\urlprefix\url{https://www.science.org/doi/10.1126/sciadv.abk0644}

\end{thebibliography}
\end{document}